%% file: main.tex
\documentclass{article}
\usepackage{microtype}
\usepackage{graphicx}
\usepackage{subfigure}
\usepackage{booktabs}
\usepackage{amssymb}
\usepackage{listings}
\usepackage{amsfonts}
\usepackage{bbm}
\usepackage{comment}
\usepackage{mathtools}

\usepackage[normalem]{ulem}
\usepackage{hyperref}
\usepackage{color}
\usepackage{setspace}

\definecolor{codeblue}{rgb}{0.25,0.5,0.5}
\definecolor{keywords}{rgb}{0.08,0.54,.02}
\usepackage[accepted]{icml2020}

\icmltitlerunning{CURL: Contrastive Unsupervised Representations for Reinforcement Learning}

\begin{document}

\twocolumn[
\icmltitle{CURL: Contrastive Unsupervised Representations for Reinforcement Learning}

\icmlsetsymbol{equal}{*}

\begin{icmlauthorlist}
\icmlauthor{Aravind Srinivas$^*$}{bair}
\icmlauthor{Michael Laskin$^*$}{bair}
\icmlauthor{Pieter Abbeel}{bair}
\end{icmlauthorlist}

\icmlaffiliation{bair}{University of California, Berkeley, BAIR}

\icmlcorrespondingauthor{Aravind Srinivas, Michael Laskin}{aravind\_srinivas, mlaskin@berkeley.edu}

\icmlkeywords{Contrastive Learning, Reinforcement Learning}
\vskip 0.3in ]

\printAffiliationsAndNotice{\icmlEqualContribution} %
\input{abstract}
\input{introduction}
\input{related}
\input{background}

\input{experiments}

\input{results}

\input{conclusion}

\newpage
\bibliography{main}
\bibliographystyle{icml2020}

\newpage
\appendix
\input{appendix}
\input{log}
\input{rad}

\end{document}

%% file: abstract.tex
\begin{abstract}
We present CURL: \textbf{C}ontrastive \textbf{U}nsupervised Representations for \textbf{R}einforcement \textbf{L}earning. CURL extracts high-level features from raw pixels using contrastive learning and performs off-policy control on top of the extracted features. CURL outperforms prior pixel-based methods, both model-based and model-free, on complex tasks in the DeepMind Control Suite and Atari Games showing 1.9x and 1.2x performance gains at the 100K environment and interaction steps benchmarks respectively. On the DeepMind Control Suite, CURL is the first image-based algorithm to nearly match the sample-efficiency of methods that use state-based features. Our code is open-sourced and available at \url{https://www.github.com/MishaLaskin/curl}.
\end{abstract}

%% file: introduction.tex
\section{Introduction}
\label{introduction}

Developing agents that can perform complex control tasks from high dimensional observations such as pixels has been possible by combining the expressive power of deep neural networks with the long-term credit assignment power of reinforcement learning algorithms. Notable successes include learning to play a diverse set of video games from raw pixels \cite{mnih2015human}, continuous control tasks such as controlling a simulated car from a dashboard camera \cite{lillicrap2015continuous} and subsequent algorithmic developments and applications to agents that successfully navigate mazes and solve complex tasks from first-person camera observations \cite{jaderberg2016reinforcement, espeholt2018impala, jaderberg2019human}; and robots that successfully grasp objects in the real world \cite{kalashnikov2018qt}.

\begin{figure}[!ht]
\vskip 0.2in
\begin{center}
\centerline{\includegraphics[width=7cm]{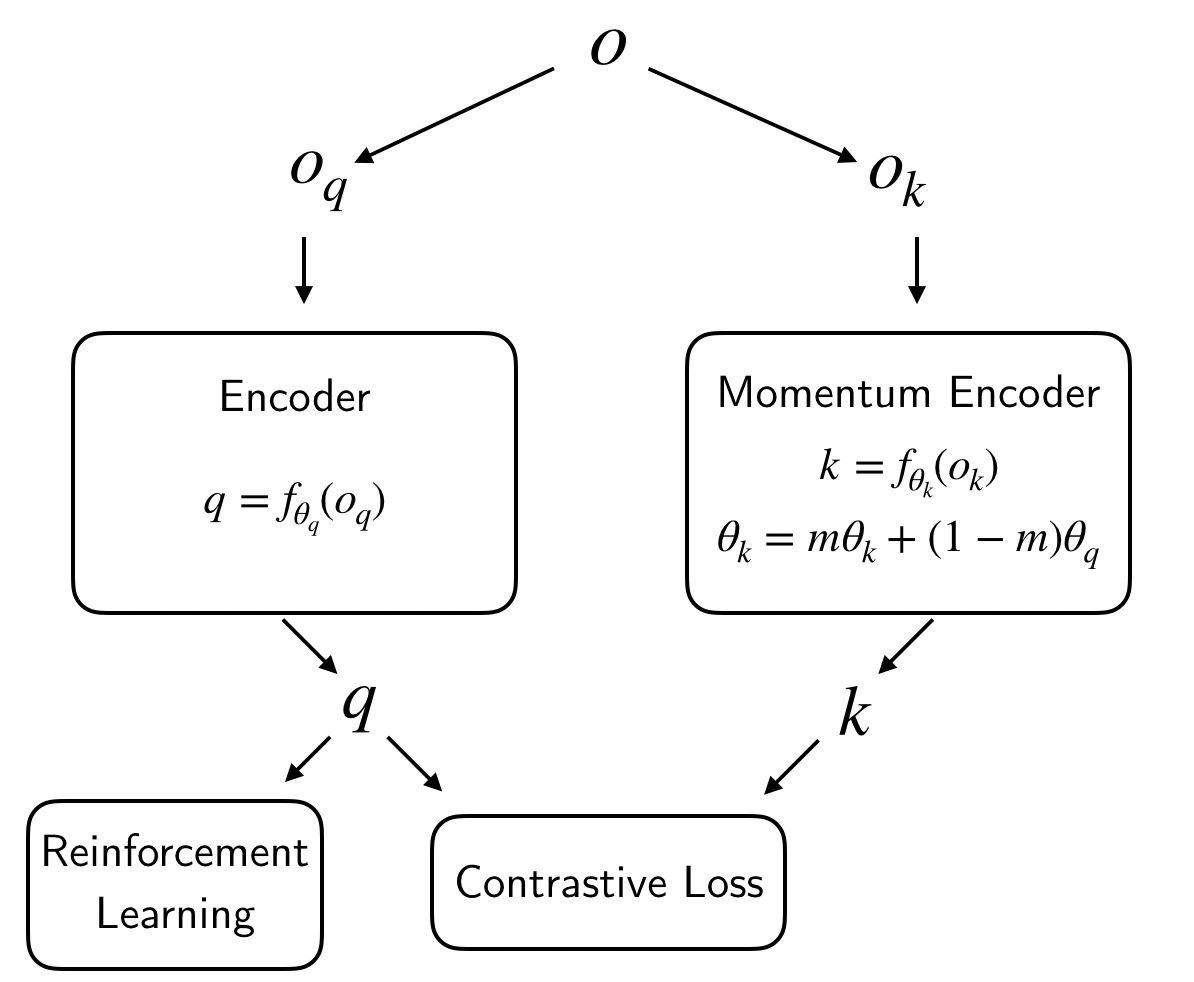}}

\caption{\textbf{C}ontrastive \textbf{U}nsupervised Representations for \textbf{R}einforcement \textbf{L}earning (CURL) combines instance contrastive learning and reinforcement learning. CURL trains a visual representation encoder by ensuring that the embeddings of data-augmented versions $o_q$ and $o_k$ of observation $o$ match using a contrastive loss. The \emph{query} observations $o_q$ are treated as the anchor while the \emph{key} observations $o_k$ contain the positive and negatives, all constructed from the minibatch sampled for the RL update. The keys are encoded with a momentum averaged version of the query encoder. The RL policy and (or) value function are built on top of the query encoder which is jointly trained with the contrastive and reinforcement learning objectives. CURL is a generic framework that can be plugged into any RL algorithm that relies on learning representations from high dimensional images.}

\label{momentum_encoder}
\end{center}
\vskip -0.2in
\end{figure}

However, it has been empirically observed that reinforcement learning from high dimensional observations such as raw pixels is sample-inefficient \cite{lake2017building, kaiser2019model}. Moreover, it is widely accepted that learning policies from physical state based features is significantly more sample-efficient than learning from pixels \cite{tassa2018deepmind}. In principle, if the state information is present in the pixel data, then we should be able to learn representations that extract the relevant state information. For this reason, it may be possible to learn from pixels as fast as from state given the right representation.

From a practical standpoint, although high rendering speeds in simulated environments enable RL agents to solve complex tasks within reasonable wall clock time, learning in the real world means that agents are bound to work within the limitations of physics. \citet{kalashnikov2018qt} needed a farm of robotic arms that collected large scale robot interaction data over several months to develop their robot grasp value functions and policies. The data-efficiency of the whole pipeline thus has significant room for improvement. Similarly, in simulated worlds which are limited by rendering speeds in the absence of GPU accelerators, data efficiency is extremely crucial to have a fast experimental turnover and iteration. Therefore, improving the sample efficiency of reinforcement learning (RL) methods that operate from high dimensional observations is of paramount importance to RL research both in simulation and the real world and allows for faster progress towards the broader goal of developing intelligent autonomous agents.

A number of approaches have been proposed in the literature to address the sample inefficiency of deep RL algorithms. Broadly, they can be classified into two streams of research, though not mutually exclusive: (i) Auxiliary tasks on the agent's sensory observations; (ii) World models that predict the future. While the former class of methods use auxiliary self-supervision tasks to accelerate the learning progress of model-free RL methods \cite{jaderberg2016reinforcement, mirowski2016learning}, the latter class of methods build explicit predictive models of the world and use those models to plan through or collect fictitious rollouts for model-free methods to learn from \cite{sutton1990integrated, ha2018world, kaiser2019model, schrittwieser2019mastering}.

Our work falls into the first class of models, which use auxiliary tasks to improve sample efficiency. Our hypothesis is simple: {\it If an agent learns a useful semantic representation from high dimensional observations, control algorithms built on top of those representations should be significantly more data-efficient.} Self-supervised representation learning has seen dramatic progress in the last couple of years with huge advances in masked language modeling \cite{devlin2018bert} and contrastive learning \cite{henaff2019data, he2019momentum, chen2020simclr} for language and vision respectively. The representations uncovered by these objectives improve the performance of any supervised learning system especially in scenarios where the amount of labeled data available for the downstream task is really low.

We take inspiration from the contrastive pre-training successes in computer vision. However, there are a couple of key differences: (i) There is no giant unlabeled dataset of millions of images available beforehand - the dataset is collected online from the agent's interactions and changes dynamically with the agent's experience; (ii) The agent has to perform unsupervised and reinforcement learning simultaneously as opposed to fine-tuning a pre-trained network for a specific downstream task. These two differences introduce a different challenge: How can we use contrastive learning for improving agents that can learn to control effectively and efficiently from online interactions?

To address this challenge, we propose CURL - \textbf{C}ontrastive \textbf{U}unsupervised Representations for \textbf{R}einforcement \textbf{L}earning. CURL uses a form of contrastive learning that maximizes agreement between augmented versions of the same observation, where each observation is a stack of temporally sequential frames. We show that CURL significantly improves sample-efficiency over prior pixel-based methods by performing contrastive learning simultaneously with an off-policy RL algorithm. CURL coupled with the Soft-Actor-Critic (SAC) \cite{haarnoja2018soft} results in {\textbf{1.9x}} median higher performance over Dreamer, a prior state-of-the-art algorithm on DMControl environments, benchmarked at {\textbf{100k}} {\it environment steps} and {\it matches the performance of state-based SAC} on the majority of 16 environments tested, a {\textbf{first}} for pixel-based methods. In the Atari setting benchmarked at 100k {\it interaction steps}, we show that CURL coupled with a data-efficient version of Rainbow DQN \cite{van2019use} results in {\textbf{1.2x}} median higher performance over prior methods such as SimPLe~\cite{kaiser2019model}, improving upon Efficient Rainbow~\cite{van2019use} on {\it 19 out of 26} Atari games, {\it surpassing human efficiency} on two games. 

While contrastive learning in aid of model-free RL has been studied in the past by \citet{oord2018representation} using Contrastive Predictive Coding (CPC), the results were mixed with marginal gains in a few DMLab \cite{espeholt2018impala} environments. CURL is the first model to show substantial data-efficiency gains from using a contrastive self-supervised learning objective for model-free RL agents across a multitude of pixel based continuous and discrete control tasks in DMControl and Atari.

We prioritize designing a simple and easily reproducible pipeline. While the promise of auxiliary tasks and learning world models for RL agents has been demonstrated in prior work, there's an added layer of complexity when introducing components like modeling the future in a latent space \cite{oord2018representation, ha2018world}. CURL is designed to add minimal overhead in terms of architecture and model learning. The contrastive learning objective in CURL operates with the same latent space and architecture typically used for model-free RL and seamlessly integrates with the training pipeline without the need to introduce multiple additional hyperparameters. 

Our paper makes the following {\textbf{key contributions}}: We present CURL, a simple framework that integrates contrastive learning with model-free RL with minimal changes to the architecture and training pipeline. Using 16 complex control tasks from the DeepMind control (DMControl) suite and 26 Atari games, we empirically show that contrastive learning combined with model-free RL outperforms the prior state-of-the-art by 1.9x on DMControl and 1.2x on Atari compared across leading prior pixel-based methods. CURL is also the first algorithm {\it across both model-based and model-free methods} that operates purely from pixels, and nearly matches the performance and sample-efficiency of a SAC algorithm trained from the state based features on the DMControl suite. Finally, our design is simple and does not require any custom architectural choices or hyperparameters which is crucial for reproducible end-to-end training.  Through these strong empirical results, we demonstrate that a contrastive objective is the preferred self-supervised auxiliary task for achieving sample-efficiency compared to reconstruction based methods, and enables {\it model-free methods to outperform state-of-the-art model-based methods in terms of data-efficiency}.

%% file: related.tex
\section{Related Work}

{\textbf{Self-Supervised Learning:}} Self-Supervised Learning is aimed at learning rich representations of high dimensional unlabeled data to be useful for a wide variety of tasks. The fields of natural language processing and computer vision have seen dramatic advances in self-supervised methods such as BERT \cite{devlin2018bert}, CPC, MoCo, SimCLR \cite{henaff2019data, he2019momentum,chen2020simclr}.

{\textbf{Contrastive Learning:}} Contrastive Learning is a framework to learn representations that obey similarity constraints in a dataset typically organized by similar and dissimilar pairs. This is often best understood as performing a dictionary lookup task wherein the positive and negatives represent a set of keys with respect to a query (or an anchor). A simple instantiation of contrastive learning is Instance Discrimination \cite{wu2018unsupervised} wherein a query and key are positive pairs if they are data-augmentations of the same instance (example, image) and negative otherwise. A key challenge in contrastive learning is the choice of negatives which can decide the quality of the underlying representations learned. The loss functions used to contrast could be among several choices such as InfoNCE \cite{oord2018representation}, Triplet \cite{wang2015unsupervised}, Siamese \cite{chopra2005learning} and so forth.

{\textbf{Self-Supervised Learning for RL}}: Auxiliary tasks such as predicting the future conditioned on the past observation(s) and action(s) \cite{jaderberg2016reinforcement, shelhamer2016loss, oord2018representation, schmidhuber1990a} are a few representative examples of using auxiliary tasks to improve the sample-efficiency of model-free RL algorithms. The future prediction is either done in a pixel space \cite{jaderberg2016reinforcement} or latent space \cite{oord2018representation}. The sample-efficiency gains from reconstruction-based auxiliary losses have been benchmarked in \citet{jaderberg2016reinforcement, higgins2017darla, yarats2019improving}. Contrastive learning has been used to extract reward signals in the latent space \cite{sermanet2018time, dwibedi2018learning, warde2018unsupervised}; and study representation learning on Atari games by \citet{anand2019unsupervised}.

{\textbf{World Models for sample-efficiency:}} While joint learning of an auxiliary unsupervised task with model-free RL is one way to improve the sample-efficiency of agents, there has also been another line of research that has tried to learn world models of the environment and use them to sample rollouts and plan. An early instantiation of the generic principle was put forth by \citet{sutton1990integrated} in Dyna where fictitious samples rolled out from a learned world model are used in addition to the agent's experience for sample-efficient learning. Planning through a learned world model~\cite{srinivas2018universal} is another way to improve sample-efficiency. While \citet{jaderberg2016reinforcement, oord2018representation, lee2019stochastic} also learn pixel and latent space forward models, the models are learned to shape the latent representations, and there is no explicit Dyna or planning.
Planning through learned world models has been successfully demonstrated in \citet{ha2018world, hafner2018learning, hafner2019dream}. \citet{kaiser2019model} introduce SimPLe which implements Dyna with expressive deep neural networks for the world model for sample-efficiency on Atari games.

{\textbf{Sample-efficient RL for image-based control:}} CURL encompasses the areas of self-supervision, contrastive learning and using auxiliary tasks for sample-efficient RL. We benchmark for sample-efficiency on the DMControl suite \citep{tassa2018deepmind} and Atari Games benchmarks \cite{bellemare2013arcade}. The DMControl suite has been used widely by \citet{yarats2019improving}, \citet{hafner2018learning}, \citet{hafner2019dream} and \citet{lee2019stochastic} for benchmarking sample-efficiency for image based continuous control methods. As for Atari, \citet{kaiser2019model} propose to use the 100k interaction steps benchmark for sample-efficiency which has been adopted in \citet{kielak2020rainbow,  van2019use}. The Rainbow DQN \cite{hessel2017rainbow} was originally proposed for maximum sample-efficiency on the Atari benchmark and in recent times has been adapted to a version known as Data-Efficient Rainbow \cite{van2019use} with competitive performance to SimPLe without learning world models. We benchmark extensively against both model-based and model-free algorithms in our experiments. For the DMControl experiments, we compare our method to Dreamer, PlaNet, SLAC, SAC+AE whereas for Atari experiments we compare to SimPLe, Rainbow, and OverTrained Rainbow (OTRainbow) and Efficient Rainbow (Eff. Rainbow).

%% file: background.tex
\vspace{-3mm}

\section{Background}

\begin{figure*}
\begin{center}
\centerline{\includegraphics[width=11cm]{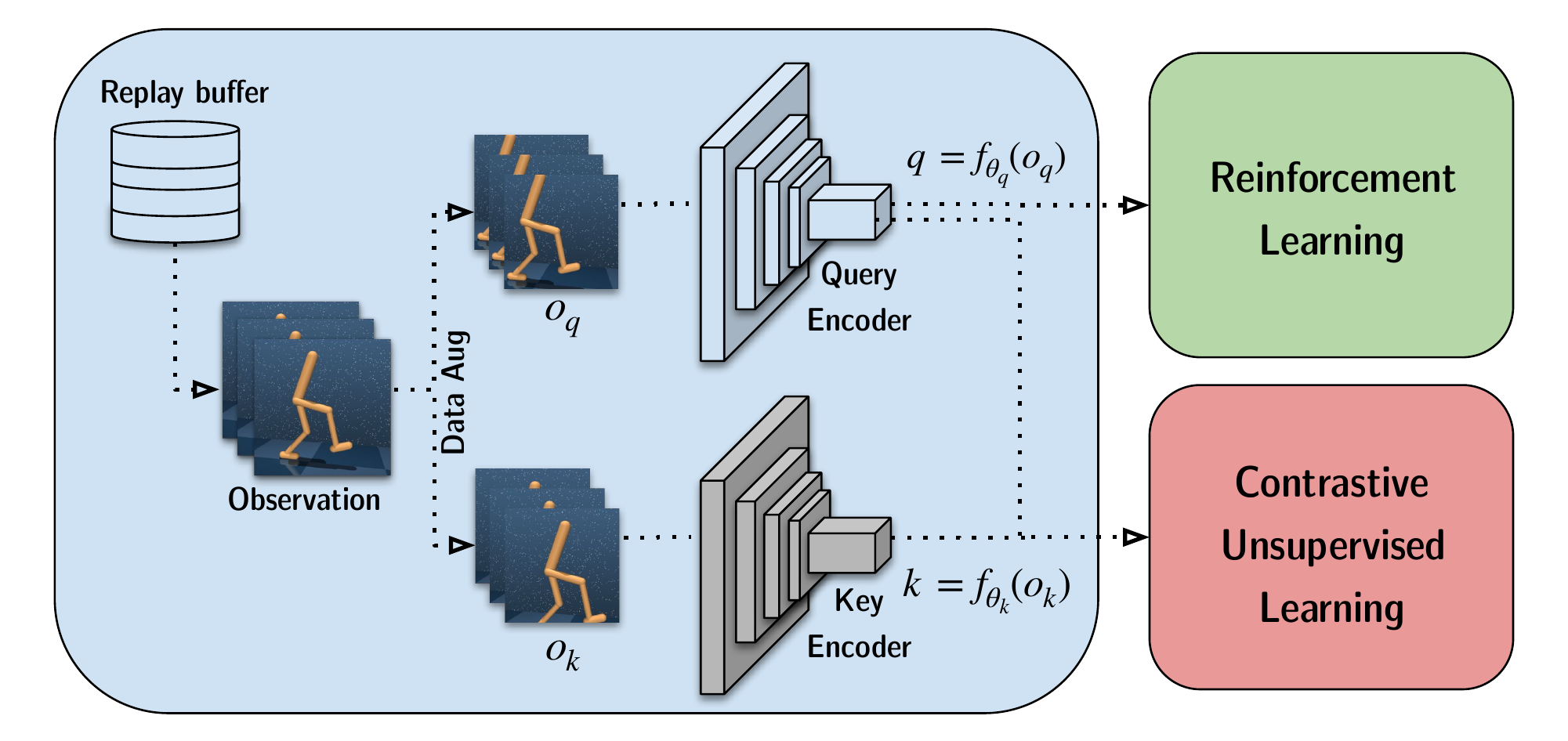}}
\caption{CURL Architecture: A batch of transitions is sampled from the replay buffer. Observations are then data-augmented twice to form {\it query} and {\it key} observations, which are then encoded with the query encoder and key encoders, respectively. The {\it queries} are passed to the RL algorithm while {\it query-key} pairs are passed to the contrastive learning objective. During the gradient update step, only the {\it query} encoder is updated. The {\it key} encoder weights are the moving average (EMA) of the query weights similar to MoCo \cite{he2019momentum}.} 
\label{fig:curl_overview_diagram}
\vspace{-10mm}
\end{center}
\end{figure*}

CURL is a general framework for combining contrastive learning with RL. In principle, one could use any RL algorithm in the CURL pipeline, be it on-policy or off-policy. We use the widely adopted Soft Actor Critic (SAC) \cite{haarnoja2018soft} for continuous control benchmarks (DM Control) and Rainbow DQN \cite{hessel2017rainbow, van2019use} for discrete control benchmarks (Atari). Below, we review SAC, Rainbow DQN and Contrastive Learning.

\subsection{Soft Actor Critic}

 SAC is an off-policy RL algorithm that optimizes a stochastic policy for maximizing the expected trajectory returns. Like other state-of-the-art end-to-end RL algorithms, SAC is effective when solving tasks from state observations but fails to learn efficient policies from pixels. SAC is an actor-critic method that learns a policy $\pi_\psi$ and critics $Q_{\phi_1}$ and $Q_{\phi_2}$. The parameters $\phi_i$ are learned by minimizing the Bellman error:
\vspace{-1mm}
\begin{equation}\label{eq:qmsbe}
   \mathcal L (\phi_{i},\mathcal{B}) = \mathbb{E}_{t \sim \mathcal B} \left [\left ( Q_{\phi_i}(o,a) - \left( r+\gamma(1-d)\mathcal T  \right )\right )^2 \right]
\end{equation}

where $t = (o,a,o',r,d)$ is a tuple with observation $o$, action $a$, reward $r$ and done signal $d$, $\mathcal{B}$ is the replay buffer, and $\mathcal T$ is the target, defined as:

\begin{equation}\label{eq:target}
   \mathcal{T}  = \left (\min_{i=1,2} Q^{*}_{\phi_i} (o',a') - \alpha \log \pi_\psi(a'|o')\right )
\end{equation}

In the target equation \eqref{eq:target}, $Q^{*}_{\phi_i}$ denotes the exponential moving average (EMA) of the parameters of $Q_{\phi_i}$. Using the EMA has empirically shown to improve training stability in off-policy RL algorithms. The parameter $\alpha$ is a positive entropy coefficient that determines the priority of the entropy maximization over value function optimization. 

While the critic is given by $Q_{\phi_{i}}$, the actor samples actions from policy $\pi_\psi$ and is trained by maximizing the expected return of its actions as in:

\begin{equation}\label{eq:actorloss}
   \mathcal L (\psi) = \mathbb{E}_{a \sim \pi} \left [ Q^\pi (o,a) - \alpha \log \pi_\psi (a|o) \right ]
\end{equation}

where actions are sampled stochastically from the policy $
   a_\psi (o,\xi) \sim \tanh \left (\mu_\psi (o) + \sigma_\psi (o) \odot \xi \right )$ and $\xi \sim \mathcal N (0,I)$ is a standard normalized noise vector. 

\subsection{Rainbow}
Rainbow DQN \cite{hessel2017rainbow} is best summarized as multiple improvements on top of the original Nature DQN \cite{mnih2015human} applied together. Specifically, Deep Q Network (DQN) \cite{mnih2015human} combines the off-policy algorithm Q-Learning with a convolutional neural network as the function approximator to map raw pixels to action value functions. Since then, multiple improvements have been proposed such as Double Q Learning \cite{van2016deep}, Dueling Network Architectures \cite{wang2015dueling}, Prioritized Experience Replay \cite{schaul2015prioritized}, and Noisy Networks \cite{fortunato2017noisy}. Additionally, distributional reinforcement learning \cite{bellemare2017distributional} proposed the technique of predicting a distribution over possible value function bins through the C51 Algorithm. Rainbow DQN combines all of the above techniques into a single off-policy algorithm for state-of-the-art sample efficiency on Atari benchmarks. Additionally, Rainbow also makes use of multi-step returns \cite{sutton1998introduction}. \citet{van2019use} propose a data-efficient version of the Rainbow which can be summarized as an improved configuration of hyperparameters that is optimized for performance benchmarked at 100K interaction steps.

\subsection{Contrastive Learning}

A key component of CURL is the ability to learn rich representations of high dimensional data using contrastive unsupervised learning. Contrastive learning \cite{hadsell2006dimensionality, lecun2006tutorial, oord2018representation, wu2018unsupervised, he2019momentum} can be understood as learning a differentiable dictionary look-up task. Given a query $q$ and keys $\mathbb{K} = \{k_0, k_1, \dots \}$ and an explicitly known partition of $\mathbb{K}$ (with respect to $q$) $P(\mathbb{K}) = ( \{k_{+}\}, \mathbb{K} \setminus \{k_{+}\} )$, the goal of contrastive learning is to ensure that $q$ matches with $k_{+}$ relatively more than any of the keys in $\mathbb{K} \setminus \{k_{+}\}$. $q, \mathbb{K}, k_{+},$ and $\mathbb{K} \setminus \{k_{+}\}$ are also referred to as {\it anchor, targets, positive, negatives} respectively in the parlance of contrastive learning \cite{oord2018representation, he2019momentum}. Similarities between the anchor and targets are best modeled with dot products ($q^Tk)$ \cite{wu2018unsupervised, he2019momentum} or bilinear products ($q^TWk$) \cite{oord2018representation, henaff2019data} though other forms like euclidean distances are also common \cite{schroff2015facenet, wang2015unsupervised}. To learn embeddings that respect these similarity relations, \citet{oord2018representation} propose the InfoNCE loss:

\begin{equation}\label{eq:infonce}
   \mathcal L_{q} = \log \frac{\exp(q^T W k_{+})}{ \exp(q^T W k_{+}) +  \sum_{i=0}^{K-1} \exp(q^T W k_{i})}
\end{equation}

The loss \ref{eq:infonce} can be interpreted as the log-loss of a $K$-way softmax classifier whose label is $k_{+}$.

\section{CURL Implementation}

CURL minimally modifies a base RL algorithm by training the contrastive objective as an auxiliary loss during the batch update. In our experiments, we train CURL alongside two model-free RL algorithms --- SAC for DMControl experiments and Rainbow DQN (data-efficient version) for Atari experiments. To specify a contrastive learning objective, we need to define (i) the discrimination objective (ii) the transformation for generating query-key observations (iii) the embedding procedure for transforming observations into queries and keys and (iv) the inner product used as a similarity measure between the query-key pairs in the contrastive loss. The exact specification these aspects largely determine the quality of the learned representations.

We first summarize the CURL architecture, and then cover each architectural choice in detail.

\subsection{Architectural Overview}

CURL uses instance discrimination with similarities to SimCLR \cite{chen2020simclr}, MoCo \cite{he2019momentum} and CPC \cite{henaff2019data}. Most Deep RL architectures operate with a stack of temporally consecutive frames as input \cite{hessel2017rainbow}. Therefore, instance discrimination is performed across the frame stacks as opposed to single image instances. We use a momentum encoding procedure for targets similar to MoCo \cite{kaiming2019moco} which we found to be better performing for RL. Finally, for the InfoNCE score function, we use a bi-linear inner product similar to CPC \cite{oord2018representation} which we found to work better than unit norm vector products used in MoCo and SimCLR. Ablations for both the encoder and the similarity measure choices are shown in Figure \ref{fig:method_ablations}. The contrastive representation is trained jointly with the RL algorithm, and the latent code receives gradients from both the contrastive objective and the Q-function. An overview of the architecture is shown in in Figure \ref{fig:curl_overview_diagram}.

\subsection{Discrimination Objective}
A key component of contrastive representation learning is the choice of positives and negative samples relative to an anchor \cite{bachman2019learning, tian2019contrastive, henaff2019data, he2019momentum, chen2020simclr}. Contrastive Predictive Coding (CPC) based pipelines \cite{henaff2019data, oord2018representation} use groups of image patches separated by a carefully chosen spatial offset for anchors and positives while the negatives come from other patches within the image and from other images. 

While patches are a powerful way to incorporate spatial and instance discrimination together, they introduce extra hyperparameters and architectural design choices which may be hard to adapt for a new problem. SimCLR \cite{chen2020simclr} and MoCo \cite{he2019momentum} opt for a simpler design where there is no patch extraction. 

Discriminating transformed image instances as opposed to image-patches within the same image optimizes a simpler instance discrimination objective \cite{wu2018unsupervised} with the InfoNCE loss and requires minimal architectural adjustments \cite{kaiming2019moco, chen2020simclr}. It is preferable to pick a simpler discrimination objective in the RL setting for two reasons. First, considering the brittleness of reinforcement learning algorithms \cite{henderson2018deep}, complex discrimination may destabilize the RL objective. Second, since RL algorithms are trained on dynamically generated datasets, a complex discrimination objective may significantly increase the wall-clock training time. CURL therefore uses instance discrimination rather than patch discrimination.  One could view contrastive instance discrimination setups like SimCLR and MoCo as maximizing mutual information between an image and its augmented version. The reader is encouraged to refer to \citet{oord2018representation, hjelm2018learning, tschannen2019mutual} for connections between contrastive learning and mutual information.

\subsection{Query-Key Pair Generation}
Similar to instance discrimination in the image setting \cite{kaiming2019moco, chen2020simclr}, the anchor and positive observations are two different augmentations of the same image while negatives come from other images. CURL primarily relies on the random crop data augmentation, where a random square patch is cropped from the original rendering.

A significant difference between RL and computer vision settings is that an instance ingested by a model-free RL algorithm that operates from pixels is not just a single image but a stack of frames \cite{mnih2015human}. For example, one typically feeds in a stack of 4 frames in Atari experiments and a stack of 3 frames in DMControl. This way, performing instance discrimination on frame stacks allows CURL to learn both spatial and temporal discriminative features. For details regarding the extent to which CURL captures temporal features, see Appendix \ref{appendix:ablation}.

\begin{figure}[ht]
\begin{center}
\centerline{\includegraphics[width=6cm]{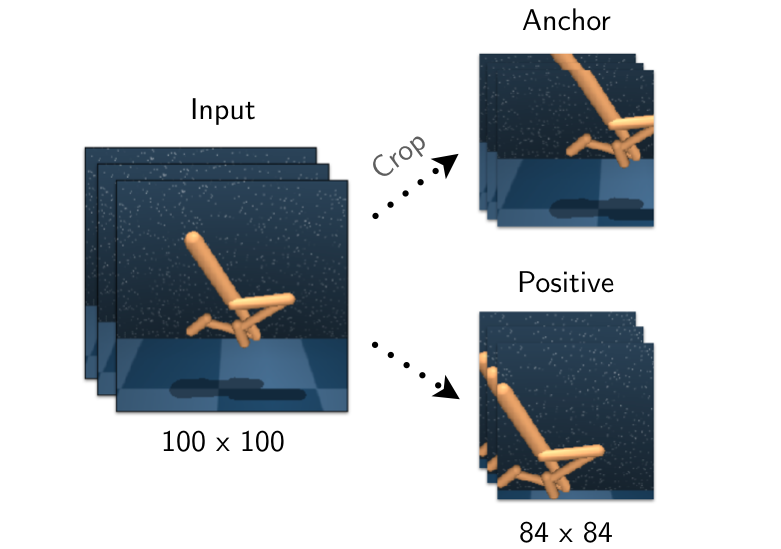}}
\caption{Visually illustrating the process of generating an anchor and its positive using stochastic random crops. Our aspect ratio for cropping is 0.84, i.e, we crop a $84 \times 84$ image from a $100 \times 100$ simulation-rendered image. Applying the same random crop coordinates across all frames in the stack ensures time-consistent spatial jittering.}
\label{fig:random_crop}
\end{center}
\vskip -0.2in
\end{figure}

We apply the random augmentations across the batch but consistently across each stack of frames to retain information about the temporal structure of the observation. The augmentation procedure is shown in Figure \ref{fig:random_crop}. For more details, refer to Appendix \ref{appendix:hyperparams}.

\vspace*{-4mm}
\subsection{Similarity Measure}

Another determining factor in the discrimination objective is the inner product used to measure agreement between query-key pairs. CURL employs the bi-linear inner-product $\text{sim}(q,k) = q^T W k$, where $W$ is a learned parameter matrix. We found this similarity measure to outperform the normalized dot-product (see Figure \ref{fig:method_ablations} in Appendix \ref{appendix:hyperparams}) used in recent state-of-the-art contrastive learning methods in computer vision like MoCo and SimCLR.

\vspace*{-2mm}
\subsection{Target Encoding with Momentum}

The motivation for using contrastive learning in CURL is to train encoders that map from high dimensional pixels to more semantic latents. InfoNCE is an unsupervised loss that learns encoders $f_q$ and $f_k$ mapping the raw anchors (query) $x_q$ and targets (keys) $x_k$ into latents $q = f_q(x_q)$ and $k = f_k(x_k)$, on which we apply the similarity dot products. It is common to share the same encoder between the anchor and target mappings, that is, to have $f_q = f_k$ \cite{oord2018representation, henaff2019data}. 

From the perspective of viewing contrastive learning as building differentiable dictionary lookups over high dimensional entities, increasing the size of the dictionary and enriching the set of negatives is helpful in learning rich representations. \citet{he2019momentum} propose momentum contrast (MoCo), which uses the exponentially moving average (momentum averaged) version of the query encoder $f_q$ for encoding the keys in $\mathbb{K}$. Given $f_q$ parametrized by $\theta_q$ and $f_k$ parametrized by $\theta_k$, MoCo performs the update $\theta_k = m \theta_k + (1 - m) \theta_q $ and encodes any target $x_k$ using $\textrm{SG}(f_k(x_k))$ [${\textrm{SG}}$ : Stop Gradient].

CURL couples frame-stack instance discrimination with momentum encoding for the targets during contrastive learning, and RL is performed on top of the encoder features.

\subsection{Differences Between CURL and Prior Contrastive Methods in RL}

\citet{oord2018representation} use Contastive Predictive Coding (CPC) as an auxiliary task wherein an LSTM operates on a latent space of a convolutional encoder; and both the CPC and A2C \cite{mnih2015human} objectives are jointly optimized. CURL avoids using pipelines that {\it predict the future} in a latent space such as \citet{oord2018representation, hafner2019dream}. In CURL, we opt for a simple instance discrimination style contrastive auxiliary task. 

\subsection{CURL Contrastive Learning Pseudocode (PyTorch-like)}\label{pytorch}

\begin{lstlisting}[language=Python]
# f_q, f_k: encoder networks for anchor
# (query) and target (keys) respectively.
# loader: minibatch sampler from ReplayBuffer
# B-batch_size, C-channels, H,W-spatial_dims
# x : shape : [B, C, H, W]
# C = c * num_frames; c=3 (R/G/B) or 1 (gray)
# m: momentum, e.g. 0.95
# z_dim: latent dimension
f_k.params = f_q.params
W = rand(z_dim, z_dim) # bilinear product.
for x in loader: # load minibatch from buffer
 x_q = aug(x) # random augmentation
 x_k = aug(x) # different random augmentation
 z_q = f_q.forward(x_q)
 z_k = f_k.forward(x_k)
 z_k = z_k.detach() # stop gradient
 proj_k = matmul(W, z_k.T) # bilinear product
 logits = matmul(z_q, proj_k) # B x B
  # subtract max from logits for stability
 logits = logits - max(logits, axis=1) 
 labels = arange(logits.shape[0])
 loss = CrossEntropyLoss(logits, labels) 
 loss.backward()
 update(f_q.params) # Adam
 update(W) # Adam
 f_k.params = m*f_k.params+(1-m)*f_q.params
\end{lstlisting}

%% file: experiments.tex
\section{Experiments}

\subsection{Evaluation}

We measure the data-efficiency and performance of our method and baselines at 100k and 500k \textit{environment steps} on DMControl and 100k \textit{interaction steps} (400k environment steps with action repeat of 4) on Atari, which we will henceforth refer to as \textbf{DMControl100k}, \textbf{DMControl500k} and \textbf{Atari100k} for clarity. While Atari100k benchmark has been common practice when  investigating data-efficiency on Atari  \cite{kaiser2019model, van2019use, kielak2020rainbow}, the DMControl benchmark was set at 500k environment steps because state-based RL approaches asymptotic performance on many environments at this point, and 100k steps to measure the speed of initial learning. A broader motivation is that while RL algorithms can achieve super-human performance on Atari games, they are still far less efficient than a human learner. Training for 100-500k environment steps corresponds to a few hours of human time. 

We evaluate (i) \textit{sample-efficiency} by measuring how many steps it takes the best performing baselines to match CURL performance at a fixed $T$ (100k or 500k) steps and (ii) \textit{performance} by measuring the ratio of the episode returns achieved by CURL versus the best performing baseline at $T$ steps. To be explicit, when we say data or sample-efficiency we're referring to (i) and when we say performance we're referring to (ii).

\begin{table*}[ht]
\caption{Scores achieved by CURL (mean \& standard deviation for 10 seeds) and baselines on DMControl500k and 1DMControl100k. CURL achieves state-of-the-art performance on the majority (\textbf{5} out of \textbf{6}) environments benchmarked on DMControl500k. These environments were selected based on availability of data from baseline methods (we run CURL experiments on 16 environments in total and show results in Figure \ref{fig:curl_all_dmc}). The baselines are PlaNet \cite{hafner2018learning}, Dreamer \cite{hafner2019dream}, SAC+AE \cite{yarats2019improving}, SLAC \cite{lee2019stochastic}, pixel-based SAC and state-based SAC \cite{haarnoja2018soft}. SLAC results were reported with one and three gradient updates per agent step, which we refer to as SLACv1 and SLACv2 respectively. We compare to SLACv1 since all other baselines and CURL only make one gradient update per agent step. We also ran CURL with three gradient updates per step and compare results to SLACv2 in Table \ref{table:three_grad}.}
\label{table:500kscores}
\vskip 0.15in
\begin{center}
\begin{small}
\begin{sc}
\begin{tabular}{lcccccccc}
\toprule
500K step scores & CURL & PlaNet & Dreamer & SAC+AE & SLACv1 & Pixel SAC &  State SAC \\
\midrule
Finger, spin    & \textbf{ 926 $\pm$ 45} & 561 $\pm$ 284 & 796 $\pm$ 183  & 884 $\pm$ 128 &   673 $\pm$ 92 & 179 $\pm$ 166  & 923 $\pm$ 21 \\
Cartpole, swingup & \textbf{841 $\pm$ 45}& 475 $\pm$ 71& 762 $\pm$ 27 & 735 $\pm$ 63 & - & 419 $\pm$ 40 &  848 $\pm$ 15 \\
Reacher, easy    & \textbf{929 $\pm$ 44}& 210 $\pm$ 390& 793 $\pm$ 164 & 627 $\pm$ 58 & - & 145 $\pm$ 30 &  923 $\pm$ 24 \\
Cheetah, run   & 518 $\pm$ 28 & 305 $\pm$ 131& 570 $\pm$ 253 & 550 $\pm$ 34 &  \textbf{640 $\pm$ 19} & 197 $\pm$ 15 & 795 $\pm$ 30 \\
Walker, walk      & \textbf{902 $\pm$ 43}& 351 $\pm$ 58 & 897 $\pm$ 49 & 847 $\pm$ 48  & 842 $\pm$ 51 & 42 $\pm$ 12 & 948 $\pm$ 54 \\
Ball in cup, catch  & \textbf{959 $\pm$ 27} & 460 $\pm$  380 & 879 $\pm$  87 & 794$\pm$  58  &  852 $\pm$  71 & 312$ \pm$  63 & 974 $\pm$ 33\\
\midrule
100K step scores  &  &  &  &  &  &  &  \\
\midrule
Finger, spin    &  \textbf{767 $\pm$ 56 } & 136 $\pm$ 216 & 341 $\pm$ 70& 740 $\pm$ 64 &  693 $\pm$ 141 & 179 $\pm$ 66 &  811$\pm$46 \\
Cartpole, swingup & \textbf{582$\pm$146}& 297$\pm$39& 326$\pm$27 & 311$\pm$11 & - & 419$\pm$40 & 835$\pm$22\\
Reacher, easy    & \textbf{538$\pm$233} & 20$\pm$50 & 314$\pm$155 & 274$\pm$14 & - & 145$\pm$30 & 746$\pm$25 \\
Cheetah, run   & 299 $\pm$48 & 138$\pm$88& 235$\pm$ 137 & 267$\pm$24 &  \textbf{319$\pm$56} & 197$\pm$15& 616$\pm$18 \\
Walker, walk      & \textbf{403$\pm$24}&  224$\pm$48& 277$\pm$12 & 394$\pm$22 & 361$\pm$73 &42$\pm$12& 891$\pm$82 \\
Ball in cup, catch  & \textbf{769 $\pm$ 43} & 0 $\pm$  0 & 246 $\pm$ 174 & 391$\pm$ 82 & 512 $\pm$ 110 & 312$\pm$ 63 & 746$\pm$91 \\ 
\bottomrule
\end{tabular}
\end{sc}
\end{small}
\end{center}
\vskip -0.1in
\end{table*}

\begin{table*}[h!]
\caption{Scores achieved by CURL (coupled with Eff. Rainbow) and baselines on Atari benchmarked at 100k time-steps (Atari100k). CURL achieves state-of-the-art performance on {\textbf{7}} out of {\textbf{26}} environments. Our baselines are SimPLe \cite{kaiser2019model}, OverTrained Rainbow (OTRainbow) \cite{kielak2020rainbow}, Data-Efficient Rainbow (Eff. Rainbow) \cite{van2019use}, Rainbow \cite{hessel2017rainbow}, Random Agent and Human Performance (Human). We see that CURL implemented on top of Eff. Rainbow improves over Eff. Rainbow on {\bf 19} out of {\bf 26} games. We also run CURL with 20 random seeds given that this benchmark is susceptible to high variance across multiple runs. We also see that CURL achieves superhuman performance on JamesBond and Krull.}

\label{table:100katariscores}
\vskip 0.15in
\begin{center}
\begin{small}
\begin{sc}
\begin{tabular}{lccccccc}
\toprule
Game & Human & Random  & Rainbow & SimPLe & OTRainbow & Eff. Rainbow  & CURL \\
\midrule

Alien & 7127.7 & 227.8 & 318.7  & 616.9 & {\bf 824.7} & 739.9 & 558.2 \\
Amidar & 1719.5 & 5.8 & 32.5 & 88.0 & 82.8 & {\bf 188.6}  & 142.1  \\
Assault & 742.0 & 222.4 & 231 &  527.2 & 351.9 & 431.2 & {\bf 600.6} \\
Asterix & 8503.3 & 210.0 & 243.6 & {\textbf{1128.3}} & 628.5 & 470.8 & 734.5 \\
Bank Heist & 753.1 & 14.2 & 15.55 & 34.2 & {\bf 182.1} & 51.0 & 131.6 \\
Battle Zone & 37187.5 & 2360.0 & 2360.0 & 5184.4 & 4060.6 & 10124.6 & {\bf 14870.0} \\
Boxing & 12.1 & 0.1 & -24.8 & {\textbf{9.1}} & 2.5 & 0.2 & 1.2 \\
Breakout & 30.5 & 1.7 & 1.2 & {\bf 16.4} & 9.84 & 1.9 & 4.9 \\
Chopper Command & 7387.8 & 811.0 & 120.0 & {\textbf{1246.9}} & 1033.33 & 861.8 & 1058.5 \\
crazy\_climber & 35829.4 & 10780.5 & 2254.5 & {\textbf{62583.6}} & 21327.8 & 16185.3 & 12146.5 \\
demon\_attack & 1971.0 & 152.1 & 163.6 & 208.1 & 711.8 & 508.0 & {\bf 817.6} \\
freeway & 29.6 & 0.0 & 0.0 & 20.3 & 25.0 & {\textbf{27.9}} & 26.7 \\
frostbite & 4334.7 & 65.2 & 60.2 & 254.7 & 231.6 & 866.8 & {\bf 1181.3} \\
gopher & 2412.5 & 257.6 & 431.2 & 771.0 & {\bf 778.0} & 349.5  & 669.3 \\
hero & 30826.4 & 1027.0 & 487 & 2656.6 & 6458.8 & {\textbf{6857.0}} & 6279.3 \\
jamesbond & 302.8 & 29.0 & 47.4 & 125.3 & 112.3 & 301.6 & {\bf 471.0} \\
kangaroo & 3035.0 & 52.0 & 0.0 & 323.1 & 605.4 & 779.3 & {\bf 872.5} \\
krull & 2665.5 & 1598.0 & 1468 & {\textbf{4539.9}} & 3277.9 & 2851.5 & 4229.6 \\
kung\_fu\_master & 22736.3 & 258.5  & 0. & {\textbf{17257.2}} & 5722.2 & 14346.1 & 14307.8 \\
ms\_pacman & 6951.6  & 307.3 & 67 & {\bf 1480.0} & 941.9 & 1204.1 & 1465.5 \\
Pong & 14.6   & -20.7 & -20.6 & {\textbf{12.8}} & 1.3 & -19.3 & -16.5 \\
Private eye & 69571.3 & 24.9 & 0 & 58.3 & 100.0 & 97.8 & {\bf 218.4} \\
Qbert & 13455.0 & 163.9 & 123.46 & {\textbf{1288.8}} & 509.3 & 1152.9 & 1042.4 \\
Road\_Runner & 7845.0 & 11.5 & 1588.46 & 5640.6 & 2696.7 & {\textbf{9600.0}} & 5661.0 \\
seaquest & 42054.7 & 68.4 & 131.69 & {\textbf{683.3}}  & 286.92 & 354.1 & 384.5 \\
Up\_n\_Down & 11693.2 & 533.4 & 504.6 & {\textbf{3350.3}} & 2847.6 & 2877.4  &  2955.2 \\

\bottomrule
\end{tabular}
\end{sc}
\end{small}
\end{center}
\vskip -0.1in
\end{table*}

\subsection{Environments}

Our primary goal for CURL is sample-efficient control from pixels that is broadly applicable across a range of environments. We benchmark the performance of CURL for both discrete and continuous control environments. Specifically, we focus on DMControl suite for continuous control tasks and the Atari Games benchmark for discrete control tasks with inputs being raw pixels rendered by the environments.

\textbf{DeepMind Control:} Recently, there have been a number of papers that have benchmarked for sample efficiency on challenging visual continuous control tasks belonging to the DMControl suite \cite{tassa2018deepmind} where the agent operates purely from pixels. The reason for operating in these environments is multi fold: (i) they present a reasonably challenging and diverse set of tasks; (ii) sample-efficiency of pure model-free RL algorithms operating from pixels on these benchmarks is poor; (iii) multiple recent efforts to improve the sample efficiency of both model-free and model-based methods on these benchmarks thereby giving us sufficient baselines to compare against; (iv) performance on the DM control suite is relevant to robot learning in real world benchmarks.

We run experiments on sixteen environments from DMControl to examine the performance of CURL on pixels relative to SAC with access to the ground truth state, shown in Figure \ref{fig:curl_all_dmc}. For more extensive benchmarking, we compare CURL to five leading pixel-based methods across the the six environments presented in \citet{yarats2019improving}: ball-in-cup, finger-spin, reacher-easy, cheetah-run, walker-walk, cartpole-swingup for benchmarking. 

{\textbf{Atari:}} Similar to DMControl sample-efficiency benchmarks, there have been a number of recent papers that have benchmarked for sample-efficiency on the Atari 2600 Games. \citet{kaiser2019model} proposed comparing various algorithms in terms of performance achieved within $100$K timesteps ($400$K frames, frame skip of $4$) of interaction with the environments (games). The method proposed by \citet{kaiser2019model} called SimPLe is a model-based RL algorithm. SimPLe is compared to a random agent, model-free Rainbow DQN \cite{hessel2017rainbow} and human performance for the same amount of interaction time. Recently, \citet{van2019use} and \citet{kielak2020rainbow} proposed data-efficient versions of Rainbow DQN which are competitive with SimPLe on the same benchmark. Given that the same benchmark has been established in multiple recent papers and that there is a human baseline to compare to, we benchmark CURL on all the 26 Atari Games (Table \ref{table:100katariscores}).

\subsection{Baselines for benchmarking sample efficiency}
\textbf{DMControl baselines:} We present a number of baselines for continuous control within the DMControl suite: (i) SAC-AE \cite{yarats2019improving} where the authors attempt to use a $\beta$-VAE \cite{higgins2017darla}, VAE \cite{kingma2013auto} and a regualrized autoencoder \citet{vincent2008extracting,ghosh2019} jointly with SAC; (ii) SLAC \cite{lee2019stochastic} which learns a latent space world model on top of VAE features \citet{ha2018world} and builds value functions on top; (iii) PlaNet and (iv) Dreamer \cite{hafner2018learning, hafner2019dream} both of which learn a latent space world model and explicitly plan through it; (v) Pixel SAC: Vanilla SAC operating purely from pixels \cite{haarnoja2018soft}. These baselines are competitive methods for benchmarking control from pixels. In addition to these, we also present the baseline State-SAC where the assumption is that the agent has access to low level state based features and does not operate from pixels. This baseline acts as an {\it oracle} in that it approximates the upper bound of how sample-efficient a pixel-based agent can get in these environments. 

\textbf{Atari baselines}: For benchmarking performance on Atari, we compare CURL to (i) SimPLe \cite{kaiser2019model}, the top performing model-based method in terms of data-efficiency on Atari and (ii) Rainbow DQN \cite{hessel2017rainbow}, a top-performing model-free baseline for Atari, (iii) OTRainbow \cite{kielak2020rainbow} which is an OverTrained version of Rainbow for data-efficiency, (iv) Efficient Rainbow \cite{van2019use} which is a modification of Rainbow hyperparameters for data-efficiency, (v) Random Agent \cite{kaiser2019model}, (vi) Human Performance \cite{kaiser2019model, van2019use}. All the baselines and our method are evaluated for performance after 100K {\it interaction steps} (400K frames with a frame skip of 4) which corresponds to roughly two hours of gameplay. These benchmarks help us understand how the state-of-the-art pixel based RL algorithms compare in terms of sample efficiency and also to human efficiency. {\textbf{Note:}} Scores for SimPLe and Human baselines have been reported differently in prior work \cite{kielak2020rainbow, van2019use}. To be rigorous, we take the {\it best} reported score for each individual game reported in prior work.

%% file: results.tex
\section{Results}
\subsection{DMControl} 
\begin{figure}[ht]
 \begin{center}
   \centerline{\includegraphics[width=\columnwidth]{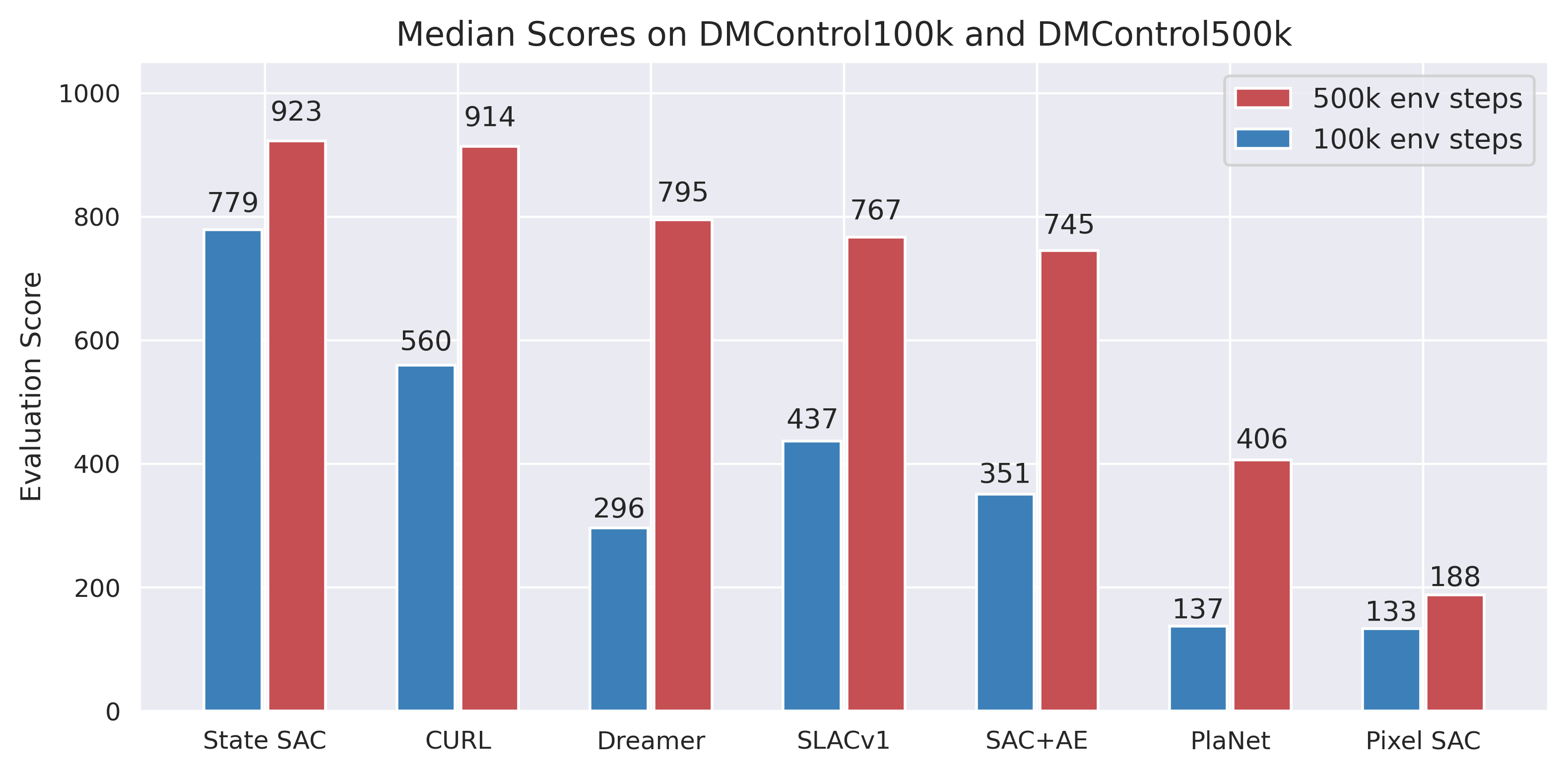}}
   \caption{Performance of CURL coupled to SAC averaged across 10 seeds relative to SLACv1, PlaNet, Pixel SAC and State SAC baselines. At the 500k benchmark CURL matches the median score of state-based SAC. At 100k environment steps CURL achieves a 1.9x higher median score than Dreamer. For a direct comparison, we only compute the median across the 6 environments in \ref{table:500kscores} (4 for SLAC) and show learning curves for CURL across 16 DMControl experiments in \ref{fig:curl_all_dmc}.} 
   \label{fig:curl_main_result}
 \end{center}
 \vskip -0.3in
 \end{figure}

Sample-efficiency results for DMControl experiments are shown in Table \ref{table:500kscores} and in Figures \ref{fig:curl_main_result}, \ref{fig:curl_sample_efficiency}, and \ref{fig:curl_all_dmc}. Below are the key findings:

(i) CURL is the \textbf{state-of-the-art image-based RL algorithm} on the majority (\textbf{5} out of \textbf{6}) DMControl environments that we benchmark on for sample-efficiency against existing pixel-based baselines. On DMControl100k, CURL achieves {\textbf{1.9x}} higher median performance than Dreamer \cite{hafner2019dream}, a leading model-based method, and is {\textbf{4.5x}} more data-efficient shown in Figure \ref{fig:curl_sample_efficiency}.

(ii) CURL operating purely from pixels {\textbf{nearly matches} (and sometimes surpasses) \textbf{the sample efficiency of SAC operating from state}} on the majority of 16 DMControl environments tested shown in Figure \ref{fig:curl_all_dmc} and matches the median state-based score on DMControl500k shown in Figure \ref{fig:curl_main_result}. This is a {\textbf{first}} for any image-based RL algorithm, be it model-based, model-free, with or without auxiliary tasks.

(iii) CURL solves (converges close to optimal score of 1000) on the majority of 16 DMControl experiments within {\textbf{500k}} steps. It also matches the state-based median score across the 6 extensively benchmarked environments in this regime.

\subsection{Atari} 
Results for Atari100k are shown in Table \ref{table:100katariscores}. Below are the key findings:

(i) CURL achieves a median human-normalized score (HNS) of {\textbf{17.5\%}} while SimPLe and Efficient Rainbow DQN achieve 14.4\% and 16.1\% respectively. The mean HNS is 38.1\%, 44.3\%, and 28.5\% for CURL, SimPLe, and Efficient Rainbow DQN respectively. 

(ii) CURL improves on top of Efficient Rainbow on {\bf 19} out of {\bf 26} Atari games. Averaged across 26 games, CURL improves on top of Efficient Rainbow by {\bf 1.3x}, while the median performance improvement over SimPLE and Efficient Rainbow are {\bf 1.2x} and {\bf 1.1x} respectively.

(iii) CURL {\textbf{surpasses human performance}} on two games JamesBond (1.6 HNS), Krull (2.5 HNS).


\section{Ablation Studies}

In Appendix \ref{appendix:ablation}, we present the results of ablation studies carried out to answer the following questions: (i) Does CURL learn only visual features or does it also capture temporal dynamics of the environment? (ii) How well does the RL policy perform if CURL representations are learned solely with the contrastive objective and no signal from RL? (iii) Why does CURL match state-based RL performance on some DMControl environments but not on others?

%% file: conclusion.tex
\section{Conclusion}

In this work, we proposed CURL, a contrastive unsupervised representation learning method for RL, that achieves state-of-the-art data-efficiency on pixel-based RL tasks across a diverse set of benchmark environments. CURL is the first model-free RL pipeline accelerated by contrastive learning with minimal architectural changes to demonstrate state-of-the-art performance on complex tasks so far dominated by approaches that have relied on learning world models and (or) decoder-based objectives. We hope that progress like CURL enables avenues for real-world deployment of RL in areas like robotics where data-efficiency is paramount.
\section{Acknowledgements} 

This research is supported in part by DARPA through the Learning with Less Labels (LwLL) Program and by ONR through PECASE N000141612723. We also thank Wendy Shang for her help with Section~\ref{noaug}; Zak Stone and Google TFRC for cloud credits; Danijar Hafner, Alex Lee, and Denis Yarats for sharing data for baselines; and Lerrel Pinto, Adam Stooke, Will Whitney, and Ankesh Anand for insightful discussions.

%% file: appendix.tex
\section{Implementation Details}
\label{appendix:hyperparams}

Below, we explain the implementation details for CURL in the DMControl setting. Specifically, we use the SAC algorithm as the RL objective coupled with CURL and build on top of the publicly released implementation from \citet{yarats2019improving}. We present in detail the hyperparameters for the architecture and optimization. We do not use any extra hyperparameter for balancing the contrastive loss and the reinforcement learning losses. Both the objectives are weighed equally in the gradient updates.

\begin{table}[h]
\caption{Hyperparameters used for DMControl CURL experiments. Most hyperparameters values are unchanged across environments with the exception for action repeat, learning rate, and batch size.}

\label{table:hyperparameters}
\vskip 0.15in
\begin{center}
\begin{small}
\begin{tabular}{ll}
\toprule
\textbf{Hyperparameter} & \textbf{Value}  \\
\midrule
Random crop    & True  \\ 
Observation rendering    & $(100,100)$  \\ 
Observation downsampling    & $(84,84)$  \\ 
Replay buffer size    & $100000$ \\ 
Initial steps    & $1000$  \\ 
Stacked frames    & $3$  \\ 
Action repeat    & $2$ finger, spin; walker, walk\\
 & $8$ cartpole, swingup \\
 & $4$ otherwise  \\
Hidden units (MLP)    & $1024$  \\ 
Evaluation episodes    & $10$  \\ 
Optimizer    & Adam  \\ 
$(\beta_1,\beta_2) \rightarrow (f_\theta, \pi_\psi, Q_\phi)$   & $(.9,.999)$  \\
$(\beta_1,\beta_2) \rightarrow (\alpha)$   & $(.5,.999)$  \\
Learning rate $(f_\theta, \pi_\psi, Q_\phi)$     & $2e-4$ cheetah, run \\
& $1e-3$ otherwise\\ 
Learning rate ($\alpha$) & $1e-4$ \\

Batch Size    & $512$  \\ 
$Q$ function EMA $\tau$ & $0.01$ \\
Critic target update freq & $2$ \\
Convolutional layers & $4$ \\
Number of filters & $32$ \\
Non-linearity & ReLU \\
Encoder EMA $\tau$ & $0.05$ \\
Latent dimension & $50$ \\
Discount $\gamma$ & $.99$ \\
Initial temperature & $0.1$ \\

\bottomrule
\end{tabular}
\end{small}
\end{center}
\vskip -0.1in
\end{table}

{\textbf{Architecture:}} We use an encoder architecture that is similar to \cite{yarats2019improving}, which we sketch in PyTorch-like pseuodocode below. The actor and critic both use the same encoder to embed image observations. A full list of hyperparameters is displayed in Table \ref{table:hyperparameters}.

\begin{figure}[H]
\vskip 0.2in
\begin{center}
\centerline{\includegraphics[width=\columnwidth]{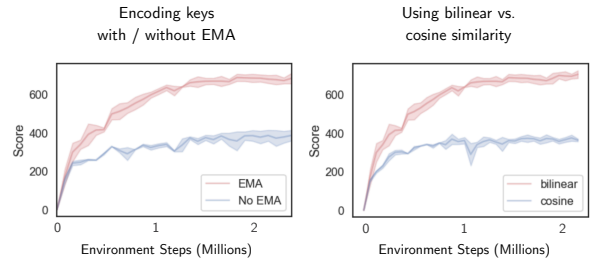}}
\caption{Performance on cheetah-run environment ablated two-ways: (left) using the query encoder or exponentially moving average of the query encoder for encoding keys (right) using the bi-linear inner product as in \cite{oord2018representation} or the cosine inner product as in \citet{kaiming2019moco,chen2020simclr}}
\label{fig:method_ablations}
\end{center}
\vskip -0.2in
\end{figure}

For contrastive learning, CURL utilizes momentum for the key encoder \cite{kaiming2019moco} and a bi-linear inner product as the similarity measure \cite{oord2018representation}. Performance curves ablating these two architectural choices are shown in Figure \ref{fig:method_ablations}.

Pseudo-code for the architecture is provided below:

\begin{lstlisting}[language=Python]
def encode(x,z_dim):
    """
    ConvNet encoder
    args:
        B-batch_size, C-channels
        H,W-spatial_dims
        x : shape : [B, C, H, W]
        C = 3 * num_frames; 3 - R/G/B
        z_dim: latent dimension
    """

    x = x / 255.
    
    # c: channels, f: filters
    # k: kernel, s: stride
    
    z = Conv2d(c=x.shape[1], f=32, k=3, s=2)])(x)
    z = ReLU(z)
    
    for _ in range(num_layers - 1):
        z = Conv2d((c=32, f=32, k=3, s=1))(z)
        z = ReLU(z)
    
    z = flatten(z)
    
    # in: input dim, out: output_dim, h: hiddens
    
    z = mlp(in=z.size(),out=z_dim,h=1024)
    z = LayerNorm(z)
    z = tanh(z)
\end{lstlisting}

{\textbf{Terminology}:} A common point of confusion is the meaning  ``training steps." We use the term \textit{environment steps} to denote the amount of times the simulator environment is stepped through and \textit{interaction steps} to denote the number of times the agent steps through its policy. The terms \textit{action repeat} or \textit{frame skip} refer to the number of times an action is repeated when it's drawn from the agent's policy. For example, if action repeat is set to 4, then 100k interaction steps is equivalent to 400k environment steps.

{\textbf{Batch Updates:}} After initializing the replay buffer with observations extracted by a random agent, we sample a batch of observations, compute the CURL objectives, and step through the optimizer. Note that since queries and keys are generated by data-augmenting an observation, we can generate arbitrarily many keys to increase the contrastive batch size without sampling any additional observations.

{\textbf{Shared Representations:}}  
The objective of performing contrastive learning together with RL is to ensure that the shared encoder learns rich features that facilitate sample efficient control. There is a subtle coincidental connection between MoCo and off-policy RL. Both the frameworks adopt the usage of a momentum averaged (EMA) version of the underlying model. In MoCo, the EMA encoder is used for encoding the keys (targets) while in off-policy RL, the EMA version of the Q-networks are used as targets in the Bellman error \cite{mnih2015human, haarnoja2018soft}. Thanks to this connection, CURL shares the convolutional encoder, momentum coefficient and EMA update between contrastive and reinforcement learning updates for the shared parameters. The MLP part of the critic that operates on top of these convolutional features has a separate momentum coefficient and update decoupled from the image encoder parameters.

{\textbf{Balancing Contrastive and RL Updates}:}
While past work has learned hyperparameters to balance the auxiliary loss coefficient or learning rate relative to the RL objective \cite{jaderberg2016reinforcement, yarats2019improving}, CURL does not need any such adjustments. We use both the contrastive and RL objectives together with equal weight and learning rate. This simplifies the training process compared to other methods, such as training a VAE jointly \cite{hafner2018learning,hafner2019dream,lee2019stochastic}, that require careful tuning of coefficients for representation learning.

{\textbf{Differences in Data Collection between Computer Vision and RL Settings}:} 
There are two key differences between contrastive learning in the computer vision and RL settings because of their different goals. Unsupervised feature learning methods built for downstream vision tasks like image classification assume a setting where there is a large static dataset of unlabeled images.  On the other hand, in RL, the dataset changes over time to account for the agent's new experiences. Secondly, the size of the memory bank of labeled images and dataset of unlabeled ones in vision-based settings are 65K and 1M (or 1B) respectively. The goal in vision-based methods is to learn from millions of unlabeled images. On the other hand, the goal in CURL is to develop sample-efficient RL algorithms. For example, to be able to solve a task within 100K timesteps (approximately 2 hours in real-time), an agent can only ingest 100K image frames.
 
 Therefore, unlike MoCo, CURL does not use a memory bank for contrastive learning. Instead, the negatives are constructed on the fly for every minibatch sampled from the agent's replay buffer for an RL update similar to SimCLR. The exact implementation is provided as a PyTorch-like code snippet in \ref{pytorch}.

{\textbf{Data Augmentation:}}

Random crop data augmentation has been crucial for the performance of deep learning based computer vision systems in object recognition, detection and segmentation \cite{krizhevsky2012, szegedy2015, cubuk2019, chen2020simclr}. However, similar augmentation methods have not seen much adoption in the field of RL even though several benchmarks use raw pixels as inputs to the model.

CURL adopts the random crop data augmentation as the stochastic data augmentation applied to a frame stack. To make it easier for the model to correlate spatio-temporal patterns in the input, we apply the same random crop (in terms of box coordinates) across all four frames in the stack as opposed to extracting different random crop positions from each frame in the stack. Further, unlike in computer vision systems where the aspect ratio for random crop is allowed to be as low as 0.08, we preserve much of the spatial information as possible and use a constant aspect ratio of 0.84 between the original and cropped. In our experiments, data augmented samples for CURL are formed by cropping $84\times84$ frames from an input frame of $100\times100$. 

{\textbf{DMControl:}} We render observations at $100 \times 100$ and randomly crop $84 \times 84$ frames. For evaluation, we render observations at $100 \times 100$ and center crop to $84 \times 84$ pixels. We found that implementing random crop efficiently was extremely important to the success of the algorithm. We provide pseudocode below:

\begin{lstlisting}[language=Python]
from skimage import view_as_windows
import numpy as np

def random_crop(imgs, out):
    """
    Vectorized random crop
    args:
        imgs: shape (B,C,H,W)
        out: output size (e.g. 84)
    """
    
    # n: batch size.
    n = imgs.shape[0]
    img_size = imgs.shape[-1] # e.g. 100
    crop_max = img_size - out
    
    imgs = np.transpose(imgs, (0, 2, 3, 1))
    
    w1 = np.random.randint(0, crop_max, n)
    h1 = np.random.randint(0, crop_max, n)
    
    # creates all sliding window
    # combinations of size (out)
    
    windows = view_as_windows(
        imgs, (1, out, out, 1))[..., 0,:,:, 0]
    
    # selects a random window 
    # for each batch element
    cropped = windows[np.arange(n), w1, h1]
    return cropped
\end{lstlisting}

\section{Atari100k Implementation Details}

The flexibility of CURL allows us to apply it to discrete control setting with minimal modifications. Similar to our rationale for picking SAC as the baseline RL algorithm to couple CURL with (for continuous control), we pick the data-efficient version of Rainbow DQN (Efficient Rainbow) \cite{van2019use} for Atari100K which performs competitively with an older version of SimPLe (most recent version has improved numbers). In order to understand {\it specifically} what the gains from CURL are without any other changes, we adopt the {\it exact} same hyperparameters specified in the paper~\cite{van2019use} (including a modified convolutional encoder that uses larger kernel size and stride of 5). We present the details in Table \ref{table:hyperparametersatari}. Similar to DMControl, the contrastive objective and the RL objective are weighted equally for learning (except for Pong, Freeway, Boxing and PrivateEye for which we used a coefficient of $0.05$ for the momentum contastive loss. On a large majority (22 out of 26) of the games, we do not use this adjustment. While it is standard practice to use the same hyperparameters for all games in Atari, papers proposing auxiliary losses have adopted a different practice of using game specific coefficients~\cite{jaderberg2016reinforcement}.). We use the Efficient Rainbow codebase from \url{https://github.com/Kaixhin/Rainbow} which has a reproduced version of \citet{van2019use}. We evaluate with $20$ random seeds and report the mean score for each game given the high variance nature of the Atari100k steps benchmark. We restrict ourselves to using grayscale renderings of image observations and use random crop of frame stack as data augmentation.

\begin{table}[h]
\caption{Hyperparameters used for Atari100K CURL experiments. Hyperparameters are unchanged across games.}

\label{table:hyperparametersatari}
\vskip 0.15in
\begin{center}
\begin{small}
\begin{tabular}{ll}
\toprule
\textbf{Hyperparameter} & \textbf{Value}  \\
\midrule
Random crop    & True  \\ 
Image size    & $(84,84)$  \\ 
Data Augmentation & Random Crop (Train) \\
Replay buffer size    & $100000$ \\ 
Training frames & $400000$ \\ 
Training steps & $100000$ \\
Frame skip & $4$ \\
Stacked frames    & $4$  \\ 
Action repeat    & $4$ \\
Replay period every & $1$ \\
Q network: channels & $32$, $64$ \\
Q network: filter size & $5\times 5, 5\times 5$ \\
Q network: stride & $5$, $5$ \\
Q network: hidden units & $256$ \\
Momentum (EMA for CURL) $\tau$ & $0.001$  \\
Non-linearity & ReLU \\
Reward Clipping   & $[-1, 1]$  \\ 
Multi step return & $20$ \\
Minimum replay size for sampling & $1600$ \\ 
Max frames per episode & $108$K \\
Update & Distributional Double Q \\
Target Network Update Period & every $2000$ updates \\
Support-of-Q-distribution & $51$ bins \\ 
Discount $\gamma$ & $0.99$ \\
Batch Size & $32$  \\
Optimizer    & Adam  \\ 
Optimizer: learning rate & $0.0001$ \\
Optimizer: $\beta1$ & $0.9$ \\
Optimizer: $\beta2$ & $0.999$ \\ 
Optimizer $\epsilon$ & $0.000015$ \\
Max gradient norm & $10$ \\ 
Exploration & Noisy Nets   \\
Noisy nets parameter & $0.1$ \\
Priority exponent & $0.5$ \\
Priority correction & $0.4 \rightarrow 1$ \\
Hardware & CPU \\
\bottomrule
\end{tabular}
\end{small}
\end{center}
\vskip -0.1in
\end{table}

\section{Benchmarking Data Efficiency}

Tables \ref{table:500kscores} and \ref{table:100katariscores} show the episode returns of DMControl100k, DMControl500k, and Atari100k across CURL and a number of pixel-based baselines. CURL outperforms all baseline pixel-based methods across experiments on both DMControl100k and DMControl500k. On Atari100k experiments, CURL coupled with Eff Rainbow outperforms the baseline on the majority of games tested (19 out of 26 games).

 \begin{figure}[h]
 \begin{center}
   \centerline{\includegraphics[width=\columnwidth]{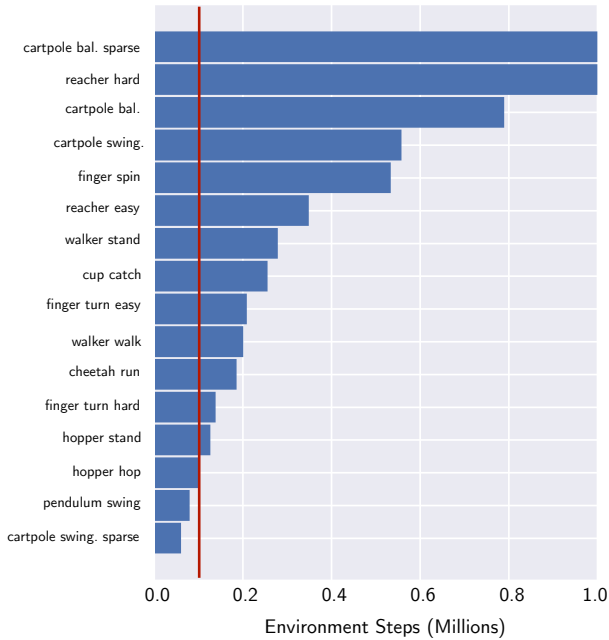}}
   \caption{The number of steps it takes a prior leading pixel-based method, Dreamer, to achieve the same score that CURL achieves at 100k training steps (clipped at 1M steps). On average, CURL is 4.5x more data-efficient. We chose Dreamer because the authors \cite{hafner2019dream} report performance for all of the above environments while other baselines like SLAC and SAC+AE only benchmark on 4 and 6 environments, respectively. For further comparison of CURL with these methods, the reader is referred to Table \ref{table:500kscores} and Figure \ref{fig:curl_main_result}.
   }
   \label{fig:curl_sample_efficiency}
 \end{center}
 \vskip -0.2in
 \end{figure}

\section{Further Investigation of Data-Efficiency in Contrastive RL}

\begin{figure}[!ht]
 \begin{center}
   \centerline{\includegraphics[width=11cm]{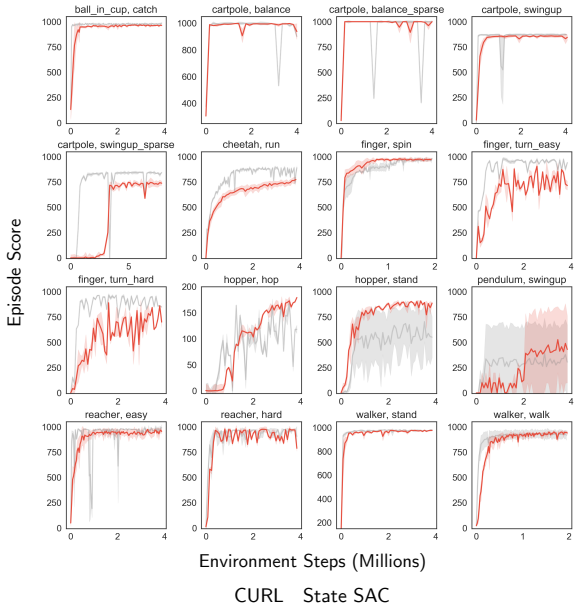}}
   \caption{CURL compared to state-based SAC run for 3 seeds on each of 16 selected DMControl environments. For the 6 environments in \ref{fig:curl_main_result}, CURL performance is averaged over 10 seeds.}
   \label{fig:curl_all_dmc}
 \end{center}
 \end{figure}

To further benchmark CURL's sample-efficiency, we compare it to state-based SAC on a total of 16 DMControl environments. Shown in Figure \ref{fig:curl_all_dmc}, CURL matches state-based data-efficiency on most of the environments, but lags behind state-based SAC on more challenging environments.

\section{Ablations}\label{appendix:ablation}

\subsection{Learning Temporal Dynamics}
\begin{figure}[t]
 \begin{center}
   \centerline{\includegraphics[width=\columnwidth]{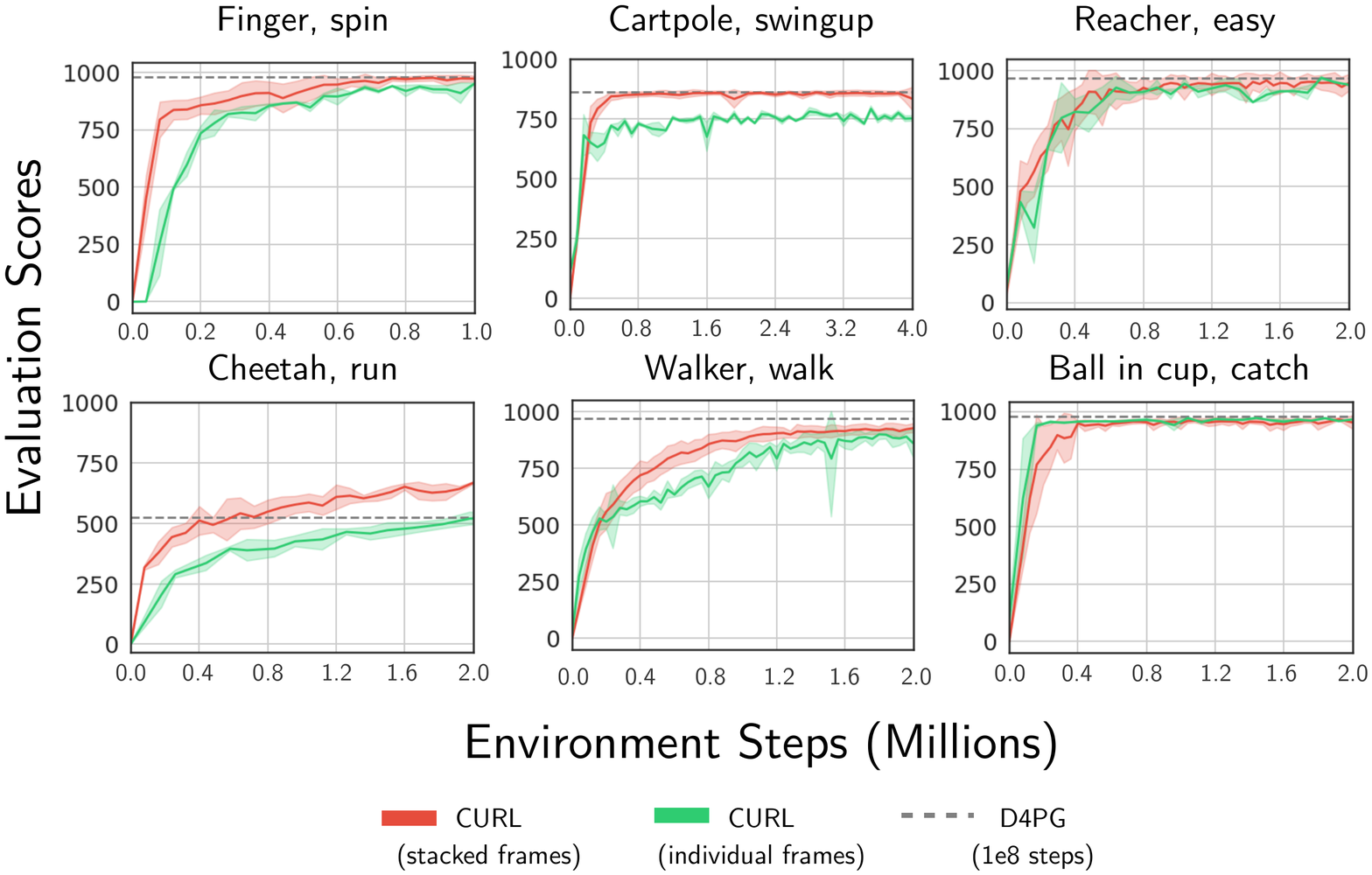}}
   \caption{CURL with temporal and visual discrimination (red) compared to CURL with only visual discrimination (green). In most settings, the variant with temporal variant outperforms the purely visual variant of CURL. The two exceptions are reacher and ball in cup environments, suggesting that learning dynamics is not necessary for those two environments. Note that the walker environment was run with action repeat of 4, whereas walker walk in the main results Table \ref{table:500kscores} and Figure \ref{fig:curl_all_dmc} was run with action repeat of 2. }
   \vspace{-6mm}
   \label{fig:curl_temporal_ablation}
 \end{center}
 \end{figure}
 
 To gain insight as to whether CURL learns temporal dynamics across the stacked frames, we also train a variant of CURL where the discriminants are individual frames as opposed to stacked ones. This can be done by sampling stacked frames from the replay buffer but only using the first frame to update the contrastive loss:

\begin{lstlisting}[language=python]
f_q = x_q[:,:3,...] # (B,C,H,W), C=9.
f_k = x_k[:,:3,...]
\end{lstlisting}

During the actor-critic update, frames in the batch are encoded individually into latent codes, which are then concatenated before being passed to a dense network.

\begin{lstlisting}[language=python]
# x: (B,C,H,W), C=9.
z1 = encode(x[:,:3,...])
z2 = encode(x[:,3:6,...])
z3 = encode(x[:,6:9,...])
z = torch.cat([z1,z2,z3],-1)
\end{lstlisting}

Encoding each frame indiviudally ensures that the contrastive objective only has access to visual discriminants. Comparing the visual and spatiotemporal variants of CURL in Figure \ref{fig:curl_temporal_ablation} shows that the variant trained on stacked frames outperforms the visual-only version in most environments. The only exceptions are reacher and ball-in-cup environments. Indeed, in those environments the visual signal is strong enough to solve the task optimally, whereas in other environments, such as walker and cheetah, where balance or coordination is required, visual information alone is insufficient.

\subsection{Increasing Gradient Updates per Agent Step}

Although most baselines we benchmark against use one gradient update per agent step, it was recently empirically shown that increasing the ratio of gradients per step improves data-efficiency in RL \cite{kielak2020rainbow}. This finding is also supported by SLAC \cite{lee2019stochastic}, where results are shown with a ratio of 1:1 (SLACv1) and 3:1 (SLACv2). We

\begin{table}[h!]
\caption{Scores achieved by CURL and SLAC when run with a 3:1 ratio of gradient updates per agent step on DMControl500k and DMControl100k. CURL achieves state-of-the-art performance on the majority (\textbf{3} out of \textbf{4}) environments on DMControl500k. Performance of both algorithms is improved relative to the 1:1 ratio reported for all baselines in Table \ref{table:500kscores} but at the cost of significant compute and wall-clock time overhead. }
\label{table:three_grad}
\vskip 0.15in
\begin{center}
\begin{small}
\begin{sc}
\begin{tabular}{lcc}
\toprule
DMControl500k & CURL & SLACv2 \\
\midrule
Finger, spin    & \textbf{923 $\pm$ 50} & 884 $\pm$ 98 \\
Walker, walk    & \textbf{911 $\pm$ 35} & 891  $\pm$ 60\\\
Cheetah, run    & 545 $\pm$ 39 & \textbf{791 $\pm$ 37}  \\
Ball in cup, catch   & \textbf{948 
$\pm$ 21 }  & 885 $\pm$ 154 \\
\midrule
DMControl100k & CURL & SLACv2 \\
\midrule
Finger, spin    & \textbf{ 741$\pm$ 118 }  & 728 $\pm$212  \\
Walker, walk    & 428 $\pm$ 59 & \textbf{513 $\pm$ 41}  \\\
Cheetah, run    & 314 $\pm$ 46 & \textbf{438 $\pm$ 76}  \\
Ball in cup, catch    & \textbf{ 899 $\pm$ 47 }& 837 $\pm$ 147  \\
\bottomrule
\end{tabular}
\end{sc}
\end{small}
\end{center}
\vskip -0.1in
\end{table}

 \subsection{Decoupling Representation Learning from Reinforcement Learning}

Typically, Deep RL representations depend almost entirely on the reward function specific to a task. However, hand-crafted representations such as the proprioceptive state are independent of the reward function. It is much more desirable to learn reward-agnostic representations, so that the same representation can be re-used across different RL tasks. We test whether CURL can learn such representations by comparing CURL to a variant where the critic gradients are backpropagated through the critic and contrastive dense feedforward networks but stopped before reaching the convolutional neural network (CNN) part of the encoder. 

\begin{figure}[!ht]
 \begin{center}
   \centerline{\includegraphics[width=\columnwidth]{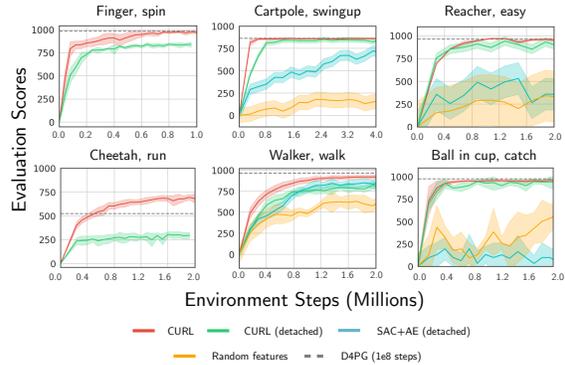}}
   \caption{CURL where the CNN part of the encoder receives gradients from both the contrastive loss and critic (red) compared to CURL with the convolutional part of the encoder trained only with the contrastive objective (green). The detached encoder variant is able to learn representations that enable near-optimal learning on most environments, except for cheetah. As in Figure \ref{fig:curl_temporal_ablation}, the walker environment was run with action repeat of 4.}
   \label{fig:detached_plots}
 \end{center}
  \vskip -.2in
  \vspace{-6mm}
 \end{figure}

Scores displayed in Figure \ref{fig:detached_plots} show that for many environments, the detached CNN representations are sufficient to learn an optimal policy. The major exception is the cheetah environment, where the detached representation significantly under-performs. Though promising, we leave further exploration of task-agnostic representations for future work.

 \subsection{Removing Data Augmentation for the Actor Critic}
 \label{noaug}

Our main results involve the use of data augmentations to regularize both the contrastive and SAC objectives. Here, we investigate whether the contrastive representations alone are sufficient for learning effective policies. In these experiments, we only augment the data for the contrastive objective but not for the SAC agent. As a result, data augmentation is used only to learn features but does not influence the control policy. The pseudocode is shown below:

\begin{lstlisting}[language=Python]
# o = original unaugmented observation
# aug = augmentation
# contrastive = InfoNCE loss
o_anchor, o_target = aug(o), aug(o)
curl_loss = contrastive(o_anchor, o_target)
sac_loss = critic_loss(o) + actor_loss(o)
loss = curl_loss + sac_loss 
params = update(params, grad(loss, params))
\end{lstlisting}

\begin{figure}[H]
\vskip 0.2in
\begin{center}
\centerline{\includegraphics[width=.7\columnwidth]{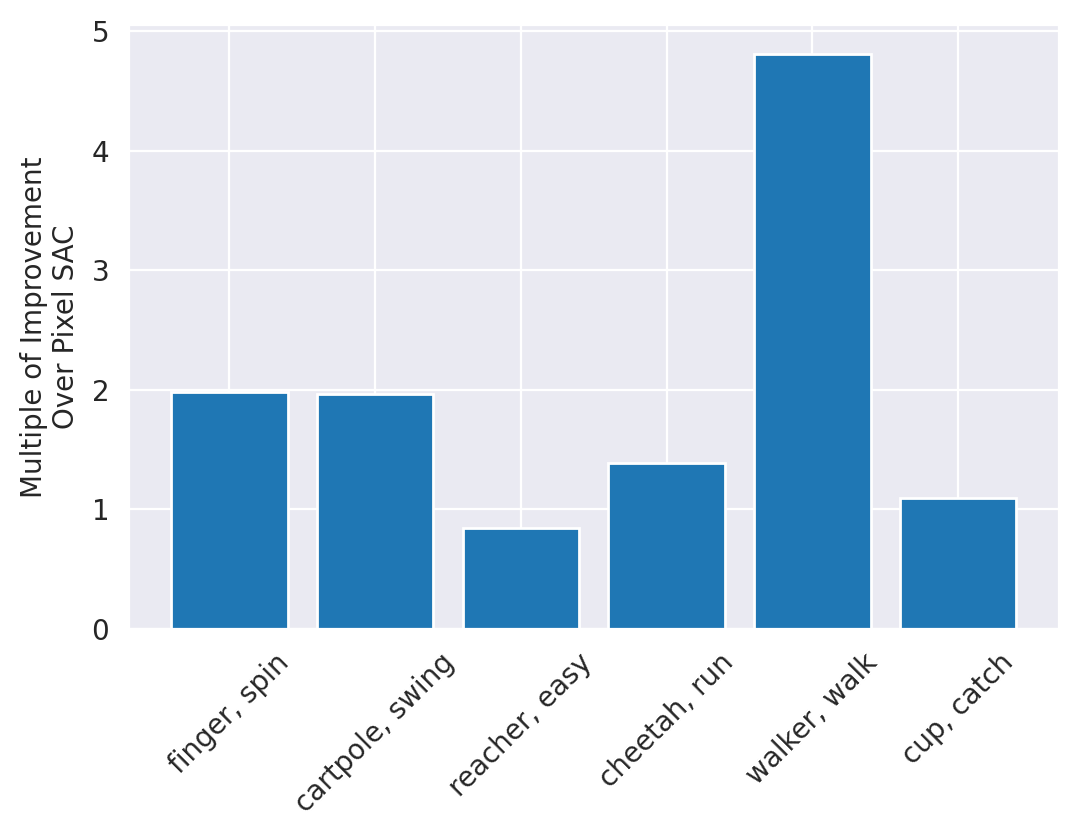}}
\caption{CURL with no data augmentations passed to the SAC agent improves the performance of the baseline pixel SAC by a mean of 2.0x / median of 1.7x on DMControl500k. For these runs we use a smaller batch size of 128 than the 512 batch size used for results in Table \ref{fig:curl_main_result}. While the constastive loss alone improves over the pixel SAC baseline, most environments benefit from data augmentation also being passed to the SAC agent.}
\label{fig:curl_noaug}
\end{center}
\vskip -0.2in
\end{figure}

DMControl500k results plotted in Figure~\ref{fig:curl_noaug} show that, on average, features learned through the contrastive loss alone improve the pixel SAC baseline by 2x. Augmenting the input passed to the SAC algorithm further improves performance.

\subsection{Predicting State from Pixels}

Despite improved sample-efficiency on most DMControl tasks, there is still a visible gap between the performance of SAC on state and SAC with CURL in some environments. Since CURL learns representations by performing instance discrimination across stacks of three frames, it's possible that the reason for degraded sample-efficiency on more challenging tasks is due to partial-observability of the ground truth state. 

\begin{figure}[!ht]
\begin{center}
\centerline{\includegraphics[width=7cm]{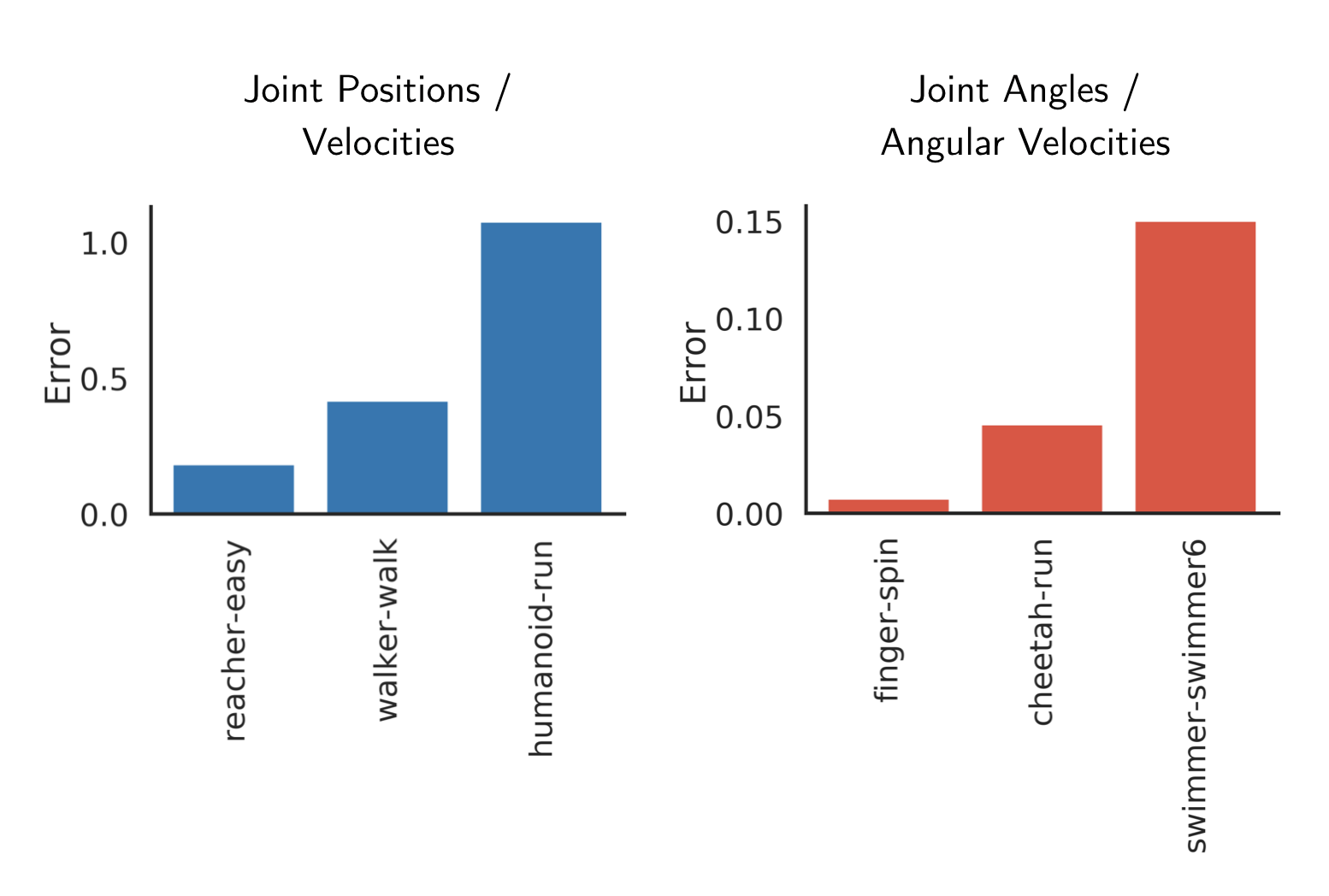}}
\caption{Test-time mean squared error for predicting the proprioceptive state from pixels on a number of DMControl environments. In DMControl, environments fall into two groups - where the state corresponds to either (a) positions and velocities of the robot joints or (b) the joint angles and angular velocities.}
   \label{fig:statereg}
\end{center}
\vskip -0.2in
\end{figure}

To test this hypothesis, we perform supervised regression $(X,Y)$ from pixels $X$ to the proprioceptive state $Y$, where each data point $x \in X$ is a stack of three consecutive frames and $y \in Y$ is the corresponding state extracted from the simulator. We find that the error in predicting the state from pixels correlates with the policy performance of pixel-based methods. Test-time error rates displayed in Figure \ref{fig:statereg} show that environments that CURL solves as efficiently as state-based SAC have low error-rates in predicting the state from stacks of pixels. The prediction error increases for more challenging environments, such as cheetah-run and walker-walk. Finally, the error is highest for environments where current pixel-based methods, CURL included, make no progress at all \cite{tassa2018deepmind}, such as humanoid and swimmer.

This investigation suggests that degraded policy performance on challenging tasks may result from the lack of requisite information about the underlying state in the pixel data used for learning representations. We leave further investigation for future work.

\subsection{CURL + Efficient Rainbow Atari runs}

We report the scores (Tables \ref{tab:atari1} and \ref{tab:atari2}) for 20 seeds across the 26 Atari games in the Atari100k benchmark for CURL coupled with Efficient Rainbow. The variance across multiple seeds is considerably high in this benchmark. Therefore, we report the scores for each of the seeds along with the mean and standard deviation for each game. 

\begin{table*}[!ht]%
\centering
\small
\setlength\tabcolsep{2.5pt}
\begin{tabular}{c|c|c|c|c|c|c|c|c|c|c|c|c}
\hline
Pacman &	Frostbite &	Asterix  &	KungFuMaster &	Kangaroo &	Gopher &	RoadRunner &	JamesBond &	BattleZone &	Seaquest &	Assault &	Krull &	Qbert			 \\
\hline
1287 &	2292 &	850 &	8470 &	600 &	1036 &	2820 &	305 &	18100 &	322 &	634.2 &	3404.3 &	1020 \\
1608 &	1046 &	525 &	10870 &	2280 &	574 &	3190 &	265 &	18200 &	236 &	696.8 &	2443.5 &	650  \\
1466 &	1209 &	655 &	10920 &	1940 &	540 &	7840 &	335 &	26800 &	352 &	655.2 &	6791.4 &	830  \\
1430 &	255 &	565 &	7730 &	1140 &	618 &	12060 &	145 &	21300 &	386 &	443 &	3022.5 &	902.5  \\
1114 &	426 &	715 &	17525 &	520 &	534 &	8340 &	565 &	7900 &	458 &	546 &	3892.2 &	3957.5  \\
1083 &	2280 &	715 &	3560 &	600 &	596 &	6920 &	565 &	8100 &	224 &	564.9 &	3505.5 &	772.5 \\
2301 &	259 &	770 &	10940 &	600 &	502 &	2230 &	350 &	12000 &	282 &	514.4 &	2564.1 &	782.5  \\
1128 &	335 &	980 &	23420 &	900 &	998 &	4250 &	365 &	16500 &	339 &	516.6 &	4079.7 &	727.5  \\
1184 &	1409 &	665 &	15160 &	600 &	950 &	1570 &	140 &	23900 &	526 &	661.5 &	2376.4 &	705 \\
1510 &	258 &	610 &	15370 &	730 &	544 &	6300 &	425 &	19900 &	436 &	664.5 &	4161.8 &	757.5  \\
2343 &	335 &	905 &	22260 &	600 &	796 &	3100 &	315 &	10000 &	272 &	529 &	3311.1 &	647.5  \\
1063 &	1062 &	800 &	17320 &	880 &	522 &	1060 &	335 &	11200 &	428 &	445.2 &	2517.3 &	562.5  \\
2040 &	1542 &	675 &	31820 &	220 &	392 &	6050 &	735 &	9700 &	358 &	573.3 &	3764.7 &	2425  \\
1195 &	1102 &	795 &	23360 &	920 &	780 &	11810 &	950 &	23500 &	533 &	531.3 &	10150.2 &	1112.5  \\
1343 &	2461 &	585	 & 27460 & 600 &	792 &	4630 &	520 &	10500 &	968 &	663.6 &	2883.6 &	527.5  \\
1354 &	257 &	865 &	7770 &	2300 &	454 &	2530 &	755 &	18100 &	314 &	795.3 &	5123.7	 & 472.5  \\
1925 &	513 &	730 &	8820 &	320 &	564 &	6840 &	750 &	9000 &	378 &	633 &	3652.5 &	610  \\
1228 &	1826 &	680 &	2980 &	600 &	522 &	6580 &	795 &	8900 &	168 &	674.1 &	2376.4 &	697.5  \\
1099 &	1889 &	965 &	10100 &	600 &	496 &	10720 &	450 &	10700 &	242 &	604.8 &	11745 &	1847.5 \\
1608 &	2869	 & 640 &	10300  & 500 &	1176 &	4380 &	355 &	13100 &	467 &	665.7 &	2826 &	840  \\
\hline
1465.5 &	1181.3 &	734.5 &	14307.8 &	872.5 &	669.3 &	5661 &	471 &	14870 &	384.5 &	600.6 &	4229.6 &	1042.4  \\ 
\hline
397.5 &	856.2 &	129.8 &	7919.3 &	600.1 &	220.6 &	3289.3 &	226.2 &	5964.3 &	170.2  & 89.5 &	2540.6 & 828.4 \\
\hline
\end{tabular}
\vspace*{-0mm}
\caption{CURL implemented on top of Efficient Rainbow - Scores reported for $20$ random seeds for each of the above games, with the last two rows being the mean and standard deviation across the runs.}
\label{tab:atari1}
\vspace{-2mm}
\end{table*}

\begin{table*}[!ht]
\centering
\small
\setlength\tabcolsep{2.5pt}
\begin{tabular}{c|c|c|c|c|c|c|c|c|c|c|c|c}
\hline
UpNDown &	Hero &	CrazyClimber &	ChopperComm. &	DemonAttack &	Amidar &	Alien &	BankHeist &	Breakout &	Freeway &	Pong &	PrivateEye &	Boxing \\
\hline
3529 & 8747.5 &	19090 &	560 &	611.5 &	150.9 &	616 &	95 &	3.6 &	29.2 &	-19.3 &	100 &	-0.5 \\
772 & 3026 &	8290 &	1530 &	707.5 &	131.2 &	923 &	184	& 5 &	25.4 &	-16.9 &	100 &	-11.4 \\
5972 & 7146 &	12160 &	1390 &	843.5 &	141.5 &	467 &	75 &	3.2 &	27.6 &	-12 &	100 &	4 \\
2793 &	7686 &	8920 &	1100 &	330.5 &	133.7 &	441 &	232 &	5.1 &	28.6 &	-19.6 &	100 &	3.6 \\
3546 &	7335 &	11360 &	500 &	759 &	157.1 &	716 &	187 &	2.9 &	22.8 &	-17.8 &	1357.4 &	6.2 \\
4552 &	7325 &	4110 &	990 &	940 &	125.4 &	453 &	367 &	6.3 &	29.6 &	-18.9 &	100 &	5 \\
2972 &	7275.5 &	9460 &	780 &	1136 &	183.2 &	273 &	186 &	5.9 &	23.3 &	-15.9 &	0 &	-1.7 \\
2865 &	3115 &	20630 &	1180 &	758 &	153.6 &	540 &	68 &	2.6 &	27.6 &	-15.2 &	100 &	0.1 \\
3098 &	7424 &	6780 &	1380 &	772.5 &	127.8 &	499 &	60 &	5.9 &	26.1 &	-18.7 &	100 &	3.5 \\
1953 &	7475 &	13570 &	970 &	820 &	149.4 &	475 &	123 &	4.3 &	28.3 &	-13.3 &	100 &	-0.5 \\
1467 &	3135 &	11890 &	1200 &	784 &	125.7 &	553 &	72 &	3.2 &	21.8 &	-17.2 &	1510 &	-22.1 \\
2912 &	5060.5 &	9160 &	1130 &	1080 &	130.4 &	446 &	53 &	4.8 &	21.8 &	-20.1 &	100 &	-1.8 \\
4123 &	4409 &	10960 &	1380 &	847 &	133 &	533 &	68 &	6.3 &	28.9 &	-16.5 &	100 &	1.6 \\
2334 &	6979 &	17360 &	1230 &	771.5 &	140.5 &	968 &	36 &	7.3 &	28.2 &	-14.9 &	100 &	3.6 \\
2605 &	4159 &	8930 &	1350 &	907.5 &	133.8 &	499 &	53 &	4.8 &	28.3 &	-19.3 &	100 &	-17.6 \\
2432 &	7560 &	11510 &	1080 &	1095.5 &	191.8 &	523 &	105 &	3.7 &	26.8 &	-15.6 &	0	 & 21.7 \\
3826 &	8587 &	22690 &	1210 &	700 &	115.5 &	616	 &276	 &6.6 &	27.5	 &-21	 &100	 &2 \\
3052 &	4683.5 &	8120 &	840 &	803.5 &	164 &	475 &	69	 &5.5 &	26.5 &	-10.5 &	0	 &5.9 \\
3131 &	7317 &	13500 &	730 &	818 &	131.7 &	525 &	50	 & 4.3	 & 26.8 &	-13.3 &	100 &	18.7 \\
1169 &	7141 &	14440 &	640 &	866 &	122.4 &	622 &	273 &	6.2 &	28.6 &	-13.1 &	100 &	3.7 \\
\hline
2955.2 & 6279.3 &	12146.5 &	1058.5 &	817.6 &	142.1 &	558.2 &	131.6 &	4.9 &	26.7 &	-16.5 &	218.4 & 1.2 \\
\hline
1181.1 & 1871.5 &	4765.6 &	299.1 &	176.6 &	20.0 &	160.3 &	94.4 &	1.4 &	2.4 & 2.9 & 417.9 &  10.0 \\
\hline
\end{tabular}
\vspace*{-0mm}
\caption{CURL implemented on top of Efficient Rainbow - Scores reported for $20$ random seeds for each of the above games, with the last two rows being the mean and standard deviation across the runs.}
\label{tab:atari2}
\vspace{-2mm}
\end{table*}

%% file: log.tex
\section{Document changelog}

This document tracks the progress and changes of CURL. In order to help readers be aware of and understand the changes, here is a brief summary:

{\bf v1} Initial version.

{\bf v2} Minor changes to DMControl to account for frame skip factor when evaluating data-efficiency of CURL and baselines. Changed action repeat for the Walker-walk task from 4 to 2 to match baseline implementations.       

{\bf v3} ICML 2020 Camera Ready. For our Atari experiments, we moved to the \url{https://github.com/Kaixhin/Rainbow} codebase for easy and clean benchmarking that directly builds on top of Efficient Rainbow without other changes. We also run 20 seeds as opposed to 3 seeds earlier given the high variance nature of the benchmark.  

{\bf v4} Added in Section~\ref{noaug} - an ablation investigating whether contrastive representations alone, with no augmentations passed to the policy during training, improve the baseline SAC policy.

%% file: rad.tex
\section{Connection to work on data augmentations}

Recently, there have been two papers published on using data augmentations for reinforcement learning, RAD~\cite{laskin2020reinforcement} and DrQ~\cite{kostrikov2020image}. These two papers present the version of CURL without an auxiliary contrastive loss but rather directly feeding in the augmented views of the image observations to the underlying value / policy network(s). Both RAD and DrQ present results on both continuous and discrete control environments, surpassing the results presented in CURL on both the DMControl and Atari benchmarks. Plenty of researchers have opined in public forums whether the results in RAD and DrQ make CURL irrelevant if the objective is to use data augmentations for data-efficient reinforcement learning. We believe that answering this question needs more nuance and present our opinions below:

1. If one has access to a rich stream of rewards from the underlying environment and is interested in optimzing the performance in terms of average reward, RAD and DrQ are likely to work better than CURL. The reason for this is simply that RAD and DrQ directly optimize for the objective one cares about, while CURL introduces an additional auxiliary consistency objective.

2. If one {\it does not} have access to a rich stream of rewards and is interested in learning good latent spaces in a {\it task agnostic manner} that can allow for data-efficient controllers across multiple tasks, CURL is the only option since the contrastive objective in CURL is reward independent. Our ablation on the detached encoder with the CURL objective present evidence that one could build simple MLPs on top of the CURL features without fine-tuning the underlying encoder and still be data-efficient on many of the DMControl tasks.

3. Future work in data-efficient reinforcement learning, particularly for real world settings, is likely to require approaches that {\it do not rely on reward functions}. In such scenarios, CURL is likely to be the more preferred approach. Further, one could potentially use CURL in a scenario where unsupervised pre-training without reward functions is initially performed before fine-tuning to the RL objective across multiple tasks.

Given the above reasons, there isn't a straightforward answer as to which is the better algorithm and the answer really depends on what the researcher / practioner wants to solve. We also emphasize that CURL was the first approach that used data augmentations effectively to significantly improve the data-efficiency of model-free reinforcement learning methods with very simple changes and showed improvement over relatively more complex model-based methods. The augmentations and results in CURL inspired future work in the form of RAD and DrQ. We hope that the analysis and results presented in CURL encourage researchers to employ data augmentations, contrastive losses and unsupervised pre-training for future reinforcement learning research.